\documentclass[10pt,twocolumn,letterpaper]{article}

\usepackage{cvpr}              
\usepackage{fp}
\usepackage{colortbl}
\usepackage{adjustbox}
\usepackage{amsfonts}
\usepackage{microtype}
\usepackage{float}
\usepackage[inline]{enumitem}
\usepackage[accsupp]{axessibility} 
\newcommand{\pad}{\phantom{0}}

\newcommand{\inlineheading}[1]{\vspace{1pt}\noindent\textbf{{#1}.}\hspace{0.5em}}

\DeclareMathAlphabet{\mymathbb}{U}{BOONDOX-ds}{m}{n}

\newcommand{\fpmin}[6]{%
  \FPmin{\result}{#2}{#3}%
  \FPmin\result{\result}{#4}%
  \FPmin\result{\result}{#5}%
  \FPmin{#1}{\result}{#6}%
}
\newcommand{\fpmax}[6]{%
  \FPmax{\result}{#2}{#3}%
  \FPmax\result{\result}{#4}%
  \FPmax\result{\result}{#5}%
  \FPmax{#1}{\result}{#6}%
}

\def\fpnorm#1#2#3{%
\newcommand{#1}[1]{%
  \FPeval{\blendfactor}{(##1 - #2)/(#3 - #2)}%
  \FPeval{\xlt}{(\blendfactor * 2.0)}%
  \FPeval{\xgt}{(\blendfactor * 2.0) - 1.0}%
  \FPeval{\ylt}{(0.75)}%
  \FPeval{\ygt}{(1.0 - 0.75)}%
  \FPeval{\biaslt}{(\xlt / ((((1.0/\ylt) - 2.0)*(1.0 - \xlt))+1.0)) * 0.5}%
  \FPeval{\biasgt}{(\xgt / ((((1.0/\ygt) - 2.0)*(1.0 - \xgt))+1.0)) * 0.5 + 0.5}%
  \FPeval{\t}{\blendfactor}%
  \FPeval{\xt}{0.5}%
  \FPiflt\t\xt \FPset{\blendfactor}{\biaslt} \else \FPset{\blendfactor}{\biasgt} \fi%
  \FPeval{\blendred}{1-(\blendfactor-0.5)/(1.0-0.5)}%
  \FPmin{\blendred}{\blendred}{1.0}%
  \FPmax{\blendred}{\blendred}{0.0}%
  \FPset{\blendgreen}{\blendred}%
  \FPset{\blendblue}{1.0}%
  \FPset{\blendalpha}{0.25}%
  \FPeval{\blendblue}{ \blendalpha * \blendblue  + (1 - \blendalpha)}%
  \FPeval{\blendgreen}{\blendalpha * \blendgreen + (1 - \blendalpha)}%
  \FPeval{\blendred}{  \blendalpha * \blendred   + (1 - \blendalpha)}%
  \xdef\temp{\noexpand\cellcolor[rgb]{\blendred, \blendgreen, \blendblue}{##1}}
  \temp
  }%
}

\def\fpnorminv#1#2#3{%
\newcommand{#1}[1]{%
  \FPeval{\blendfactor}{1 - (##1 - #2)/(#3 - #2)}%
  \FPeval{\xlt}{(\blendfactor * 2.0)}%
  \FPeval{\xgt}{(\blendfactor * 2.0) - 1.0}%
  \FPeval{\ylt}{(0.85)}%
  \FPeval{\ygt}{(1.0 - 0.85)}%
  \FPeval{\biaslt}{(\xlt / ((((1.0/\ylt) - 2.0)*(1.0 - \xlt))+1.0)) * 0.5}%
  \FPeval{\biasgt}{(\xgt / ((((1.0/\ygt) - 2.0)*(1.0 - \xgt))+1.0)) * 0.5 + 0.5}%
  \FPeval{\t}{\blendfactor}%
  \FPeval{\xt}{0.5}%
  \FPiflt\t\xt \FPset{\blendfactor}{\biaslt} \else \FPset{\blendfactor}{\biasgt} \fi%
  \FPeval{\blendred}{1-(\blendfactor-0.5)/(1.0-0.5)}%
  \FPmin{\blendred}{\blendred}{1.0}%
  \FPmax{\blendred}{\blendred}{0.0}%
  \FPset{\blendgreen}{\blendred}%
  \FPset{\blendblue}{1.0}%
  \FPset{\blendalpha}{0.25}%
  \FPeval{\blendblue}{ \blendalpha * \blendblue  + (1 - \blendalpha)}%
  \FPeval{\blendgreen}{\blendalpha * \blendgreen + (1 - \blendalpha)}%
  \FPeval{\blendred}{  \blendalpha * \blendred   + (1 - \blendalpha)}%
  \xdef\temp{\noexpand\cellcolor[rgb]{\blendred, \blendgreen, \blendblue}{##1}}
  \temp
}%
}

\newcommand{\mynormcolor}[2]{%
\FPeval{\nc}{}%

}

\definecolor{cvprblue}{rgb}{0.21,0.49,0.74}
\usepackage[pagebackref,breaklinks,colorlinks,allcolors=cvprblue]{hyperref}

\def\paperID{10515} 
\def\confName{CVPR}
\def\confYear{2025}

\title{{Geometry-guided Online 3D Video Synthesis\\with Multi-View Temporal Consistency}}

\author{Hyunho Ha\textsuperscript{1}\thanks{Work done during an internship at Meta Reality Labs.} \quad
Lei Xiao\textsuperscript{2} \quad
Christian Richardt\textsuperscript{2} \quad
Thu Nguyen-Phuoc\textsuperscript{2}\\[0.25em] 
Changil Kim\textsuperscript{2} \quad
Min H. Kim\textsuperscript{1} \quad
Douglas Lanman\textsuperscript{2} \quad
Numair Khan\textsuperscript{2}\\[0.5em]
\textsuperscript{1}KAIST \quad
\textsuperscript{2}Meta}

\begin{document}
\twocolumn[{%
\renewcommand\twocolumn[1][]{#1}%
\maketitle
\begin{center}
   \centering
   \vspace{-8mm}
   \captionsetup{type=figure}
    \includegraphics[trim=0cm 9.5cm 0.0cm 0cm, clip=true, width=0.99\linewidth]{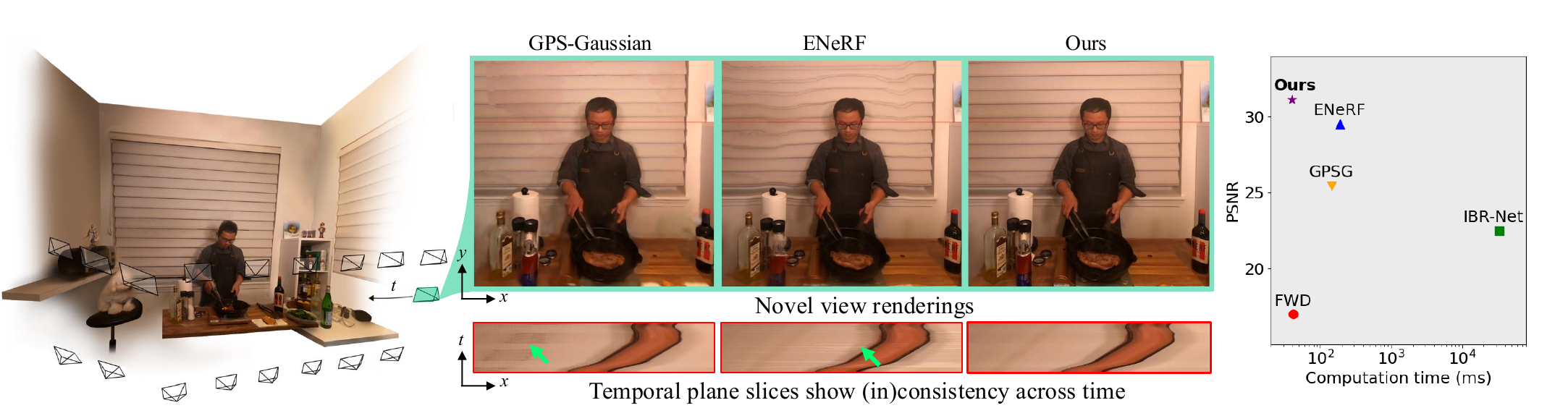}
    \captionof{figure}{
    Our method enables efficient rendering of high-quality, consistent 3D videos by addressing the challenges of view synthesis in both spatial and temporal dimensions. 
In multi-camera setups, novel-view synthesis must manage continuous changes across view and time to ensure smooth visualization. 
Existing methods process frames independently and often introduce flickering artifacts. We propose using aggregated depth as geometric guidance to a blending network to maintain consistent color and detail across frames. 
This results in view and temporally consistent 3D videos, achieving state-of-the-art quality while providing improved processing speed over prior methods.
    }
\end{center}%
}]
\renewcommand{\thefootnote}{\fnsymbol{footnote}}
\footnotetext[1]{Work done during an internship at Meta Reality Labs.} 
\footnotetext[0]{Project page: \href{https://nkhan2.github.io/projects/geometry-guided-2025}{https://nkhan2.github.io/projects/geometry-guided-2025}}
\begin{abstract}
We introduce a novel geometry-guided online video view synthesis method with enhanced view and temporal consistency. Traditional approaches achieve high-quality synthesis from dense multi-view camera setups but require significant computational resources. In contrast, selective-input methods reduce this cost but often compromise quality, leading to multi-view and temporal inconsistencies such as flickering artifacts. Our method addresses this challenge to deliver efficient, high-quality novel-view synthesis with view and temporal consistency. The key innovation of our approach lies in using global geometry to guide an image-based rendering pipeline. To accomplish this, we progressively refine depth maps using color difference masks across time. These depth maps are then accumulated through truncated signed distance fields in the synthesized view's image space. This depth representation is view and temporally consistent, and is used to guide a pre-trained blending network that fuses multiple forward-rendered input-view images. Thus, the network is encouraged to output geometrically consistent synthesis results across multiple views and time. Our approach achieves consistent, high-quality video synthesis, while running efficiently in an online manner.
\end{abstract}
\section{Introduction}
\label{sec:intro}
Recent strides in immersive 3D video are poised to revolutionize the way we experience education, video conferencing \cite{lawrence2021project, tu2024tele}, and entertainment \cite{lin2023high, xu20244k4d, JiangSHGWZYX2024} with virtual reality (VR), augmented reality (AR), and 3D video\-graphy \cite{gao2021dynamic, wang2024shape} becoming increasingly accessible and affordable. As these technologies continue to evolve, we can expect a future where immersive video content seamlessly integrates into our daily lives, transforming the way we learn, entertain, and connect with each other.

Generating accurate scene-scale reconstructions for immersive 3D video introduces unique challenges due to complex occlusions, large depth ranges, and diverse object appearances. While systems using large camera arrays have demonstrated high-quality novel-view synthesis results \cite{wilburn2005high, wilburn2004high}, processing all inputs of a dense camera rig simultaneously imposes substantial computational costs. This makes them impractical for real-time applications which require efficient processing of streaming input. This inherent trade-off between computational complexity, and camera density has spurred recent efforts to achieve high-quality view-synthesis results using sparser, more practical capture systems.

Among these, one line of work uses a fixed set of very sparsely distributed  cameras \cite{dou2016fusion4d, dou2017motion2fusion, lawrence2021project, tu2024tele}. These methods often rely on depth from additional sensors to overcome Nyquist sampling limits \cite{chai2000plenoptic, mildenhall2019local}, and provide a basic geometric framework to support appearance optimization. While such a global geometry approach ensures consistent reconstruction across views and time, it fails to represent fine detail and complex appearance changes in a scene. Furthermore, the low resolution of depth sensors limits the achievable quality of these methods in scene-scale settings, restricting them exclusively to the reconstruction of human subjects. 

Therefore, a second line of work selects a subset of optimal views from a relatively larger camera array to achieve broader scene coverage, and reconstruct more complex appearance details using an image-based rendering approach \cite{lin2022efficient, zheng2024gps, mildenhall2019local}. Further, these methods can use correspondence-based techniques to estimate high-resolution depth (explicitly, or implicitly), and overcome the Nyquist sampling limits for view synthesis required by traditional image-based rendering \cite{wilburn2005high, wilburn2004high}. Thus, recent work uses neural volumes \cite{lin2022efficient}, neural points \cite{cao2022fwd}, learned image-based rendering \cite{wang2021ibrnet}, and 3D Gaussian splats \cite{zheng2024gps} to reconstruct diverse objects' appearance. However, these methods introduce new challenges including shifts in appearance with viewpoint, and temporal inconsistencies due to lack of continuity across frames. This makes it difficult to achieve smooth reconstructions for immersive 3D video in online systems.

In this paper, we present a novel geometry-guided 3D video synthesis method that seeks to overcome these limitations by combining global geometry with an image-based rendering approach for high-quality, view and temporally consistent synthesis in dynamic scenes.

The main contribution of our method is a geometry-guided 2D image blending workflow that suppresses inconsistencies across forward-rendered images. Our approach aggregates high-resolution depth maps into a globally coherent geometric structure via a truncated signed distance field (TSDF) in the synthesized view's image space. This is used to support stable synthesis across viewpoints and time by guiding a pre-trained image-based blending network. This process ensures view-consistent and temporally coherent synthesis across diverse scene conditions, as demonstrated by results on both indoor and outdoor multi-view datasets.
\section{Related Work}
\label{sec:related-work}

\begin{figure*}[t]
  \centering
  \includegraphics[trim=0cm 8.7cm 1.5cm 0cm, clip=true, width=0.88\linewidth]{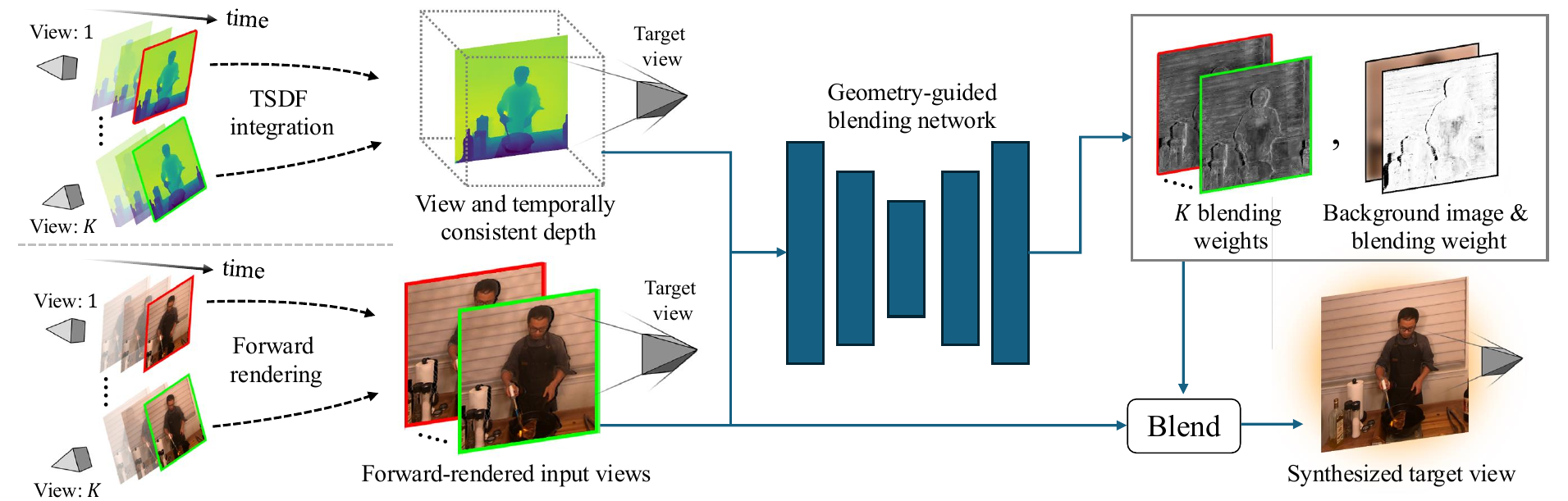}  
  \caption{\textbf{Pipeline overview.}
  Given multi-view RGB-D videos, our method forward-renders a subset of views into the target camera using 3D Gaussian splatting (\cref{subsec:pixelgs}).
  The per-view depth maps are fused using a truncated signed distance field (TSDF) that is regularized to be view and temporally consistent (\cref{subsec:tsdf}).
  This geometric guidance enables a CNN to blend the forward-rendered images and inpaint disoccluded regions, to produce consistent novel view results (\cref{subsec:blending}).
  }
\label{fig:pipeline}%
\vspace{-1em}%
\end{figure*}

\inlineheading{Voxel-based View Synthesis}
Real-time RGB-D 3D scanning methods \cite{newcombe2011kinectfusion, niessner2013real, lee2020texturefusion, ha2021normalfusion, kim2022textureme, dai2017bundlefusion} utilize the truncated signed distance field (TSDF) \cite{curless1996volumetric} to integrate geometry and color information.
These methods suppress noisy depth information but fail to handle dynamic scenes.
To address this, 4D scanning methods \cite{newcombe2015dynamicfusion, innmann2016volumedeform, slavcheva2017killingfusion, ha2020progressive, guo2017real, dou2016fusion4d, dou2017motion2fusion} estimate the motion of objects and integrate information into a canonical TSDF space.
However, they need to trade off between the resolution of expensive 3D voxel grids and computational complexity.
Recently, \citet{lawrence2021project} proposed an image-based TSDF that does not require an explicit 3D voxel grid to generate novel views, but the reconstructions are primarily focused on humans and not general dynamic scenes.
The above methods have limitations, including the need for additional depth sensors, and simple appearance models, such as diffuse or specular BRDFs.
Our algorithm overcomes these limitations by leveraging image-based rendering, enabling high-quality reconstruction of entire indoor environments with diverse moving objects.

\inlineheading{Neural-based View Synthesis}
Image-based rendering \cite{hedman2016scalable, HedmaPPFDB2018, penner2017soft} and neural-based multiplane images (MPIs) \cite{mildenhall2019local, flynn2019deepview, li2020crowdsampling, srinivasan2019pushing, zhou2018stereo, wizadwongsa2021nex, zhang2023structural} have been widely used for novel-view synthesis in static scenes, with recent applications extending to dynamic scenes \cite{lin2021deep}.
However, these methods require significant computation time to build explicit geometry or MPIs.
Volume-based neural rendering algorithms can handle more free-viewpoint rendering in 3D \cite{mildenhall2021nerf,BarroMTHMS2021, BarroMVSH2023, muller2022instant, WanRBLRNXLZRL2023, XuALGBKRPKBLZR2023} and further in 4D \cite{AttalHRZKOK2023, cao2023hexplane, fridovich2023k, li2022neural,lin2023high,xu20244k4d}, but they remain unsuitable for online use due to the high computational cost of per-scene training.
Various methods have been developed to handle generalized scene representations based on volumetric \cite{yu2021pixelnerf,xu2024murf}, depth-based \cite{cao2022fwd} or image-based rendering \cite{wang2021ibrnet}.
However, these methods fall short in handling dynamic scenes.
\citet{lin2022efficient} employ a generalized 3D CNN model to reconstruct dynamic scenes, while \citet{xiao2022neuralpassthrough} address novel-view synthesis in specific VR hardware.
However, these approaches do not explicitly address temporal and view inconsistencies.
Our algorithm does not require per-scene optimization and ensures globally and temporally consistent results across time and views by leveraging a global geometry representation.

\inlineheading{Point-based View Synthesis}
To represent diverse scene appearance, neural features \cite{aliev2020neural, rakhimov2022npbg, RueckeFS2022} are sometimes combined with 3D points.
Recently, 3D Gaussians \cite{kerbl20233d} have enabled real-time rendering for novel-view synthesis.
This approach has been extended to reconstruct dynamic environments by considering the temporal domain \cite{yang2024deformable, luiten2024dynamic, wu20244d, li2024spacetime, sun20243dgstream, li2024st, liang2023gaufre, lee2024ex4dgs}.
However, these methods require expensive per-scene optimization.
In contrast, feed-forward models \cite{chen2024mvsplat,charatan2024pixelsplat, zhu2024fsgs, chen2024lara, szymanowicz2024splatter, zhang2024gslrm, kumar2024few} have been proposed to enable novel-view synthesis in sparse static scenes.
Feed-forward Gaussian models have also been extended to reconstruct dynamic humans \cite{zheng2024gps, tu2024tele}.
Recently, \citet{guo2024real} proposed a depth-based free-viewpoint system for dynamic indoor scenes.
However, all these methods fail to explicitly account for noisy input depth maps, which results in temporal and view inconsistencies.
Our approach can handle objects with diverse appearance while maintaining consistency across view and time domains by incorporating global geometry information in 2D image-based rendering.

\section{Method}
\label{sec:method}

At a high level, our method uses depth to forward-render a subset of input images into the novel camera, and then blends these renderings to generate the novel view result. This process is repeated for each frame of a multi-camera video sequence. The individual components of our pipeline are chosen to ensure view and temporal consistency, efficient computation and state-of-the-art reconstruction quality. 

Specifically, we use pixel-sized 3D Gaussians \cite{kerbl20233d} to render the input images into the target view (\cref{subsec:pixelgs}). This allows us to handle occlusions, and avoids the aliasing artifacts of conventional point-splatting. Simultaneously, we use an image-based truncated signed distance field (TSDF) \cite{lawrence2021project} to estimate a depth map for the target view using the input-view depths (\cref{subsec:tsdf}). This TSDF depth is used to guide a blending network that generates a fused novel view from the forward-rendered images, including inpainted disoccluded regions (\cref{subsec:blending}). Using a global geometric structure like a TSDF to guide the blending network ensures the fused output remains view-consistent at any single time frame. To further ensure temporal consistency as well, we fuse the depth of static regions from previous time frames into the current frame's TSDF. Thus, our insight is to use the TSDF as a view-temporal filter between the inconsistent input depths and the synthesized novel views.

An overview of our pipeline is provided in \Cref{fig:pipeline}. At each time step $t$, the input to our method is a set of $H\!\times\!W$ multi-view RGB images $\{I_n^t, n\!=\!1, ..., N\}$ from $N$ cameras. The pose of each camera remains fixed across time, and we compute it once using COLMAP \cite{schoenberger2016sfm}. 
Since the camera configuration is known beforehand, we use a stereo algorithm \cite{lipson2021raft} to estimate depth maps $\{D_n^t,~n\!=\!1, ..., N\}$ for the input views. This also allows us to evaluate our method on existing multi-view datasets. However, it is also possible to use RGB-D cameras for input \cite{orts2016holoportation}. The output of our method is an RGB image $\mathcal{I}^t$ at the target novel view. 

\subsection{Forward Warping with 3D Gaussians}
\label{subsec:pixelgs}

We render a subset $\{I^t_k,~k\!\in\!(1, ..., N)\}$ of the $K$ closest input views into the novel camera using pixel-wise 3D Gaussians. Specifically, for each pixel of an input $I_k^t$, we use the depth map $D_k^t$ and pose to determine the 3D position of an isotropic, fully-opaque Gaussian. The RGB color of each Gaussian is set to the underlying pixel value, and its scale is computed such that its projection in the source image $I_k^t$ covers a single pixel. This step produces a set of forward-rendered images $\{\mathcal{I}_k^t\}$ along with accumulated opacity of the Gaussians $\{\alpha_k^t\}$, and depths $\{\mathcal{D}_k^t\}$ in the novel view.

\begin{figure}
\centering
\includegraphics[trim=0cm 12.7cm 19cm 0cm, clip=true, width=0.95\columnwidth]{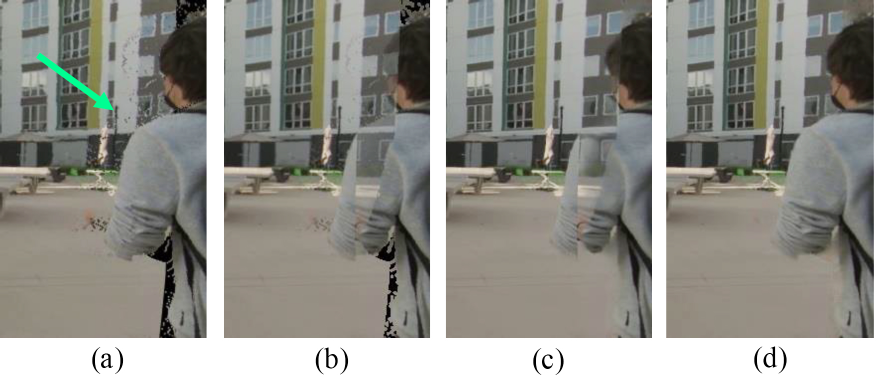}
\caption{\textbf{(a)} Rendering all input-view Gaussians into the target camera in a single pass \cite{zheng2024gps} creates flying pixels, and disocclusion holes. \textbf{(b)} On the other hand, blending multiple forward-rendered images using distance-based weights \cite{mildenhall2019local} leads to ghosting artifacts. \textbf{(c)} A naive blending network is unable to fix the ghosting. \textbf{(d)} Our method uses the target view's geometry -- rendered as a depth map from a TSDF -- to guide the blending network. This allows it to correctly fuse unstable forward-rendered input views. }
\label{fig:ablations}%
\vspace{-1em}%
\end{figure}

Our approach differs from previous methods that use feed-forward networks to predict per-pixel Gaussian parameters \cite{zheng2024gps, tu2024tele, zhang2024gslrm}. We observed that the scale, opacity and rotation parameters predicted by these networks for Gaussians on a dense $H\!\times\!W$ pixel grid converge to near-constant values, and the predicted RGB color matches the input image. Therefore, it is more efficient to compute these parameters analytically than with a neural network. With the 3D position of each Gaussian coming from the depth map, our approach achieves high-quality rendering results that match the feed-forward models in most regions.
Compared to conventional point splatting \cite{aliev2020neural}, our approach shows fewer aliasing artifacts like gaps between pixels and jagged edges. Nonetheless, disocclusions and errors in the depth maps $\{D_k^t\}$ lead to holes and flying pixels in the forward-rendered images $\{\mathcal{I}_k^t\}$. Rendering the Gaussians from all $\{I_k^t\}$ in a single pass, as \citet{zhang2024gslrm} do, fails to address these artifacts and, in fact, further deteriorates edge quality due to Z-fighting (see \cref{fig:ablations}). Instead, we blend the separate images $\{\mathcal{I}_k^t\}$ guided by the novel view's depth.

\begin{figure}
\centering
\includegraphics[trim=0cm 0cm 11cm 0cm, clip=true, width=0.95\linewidth]{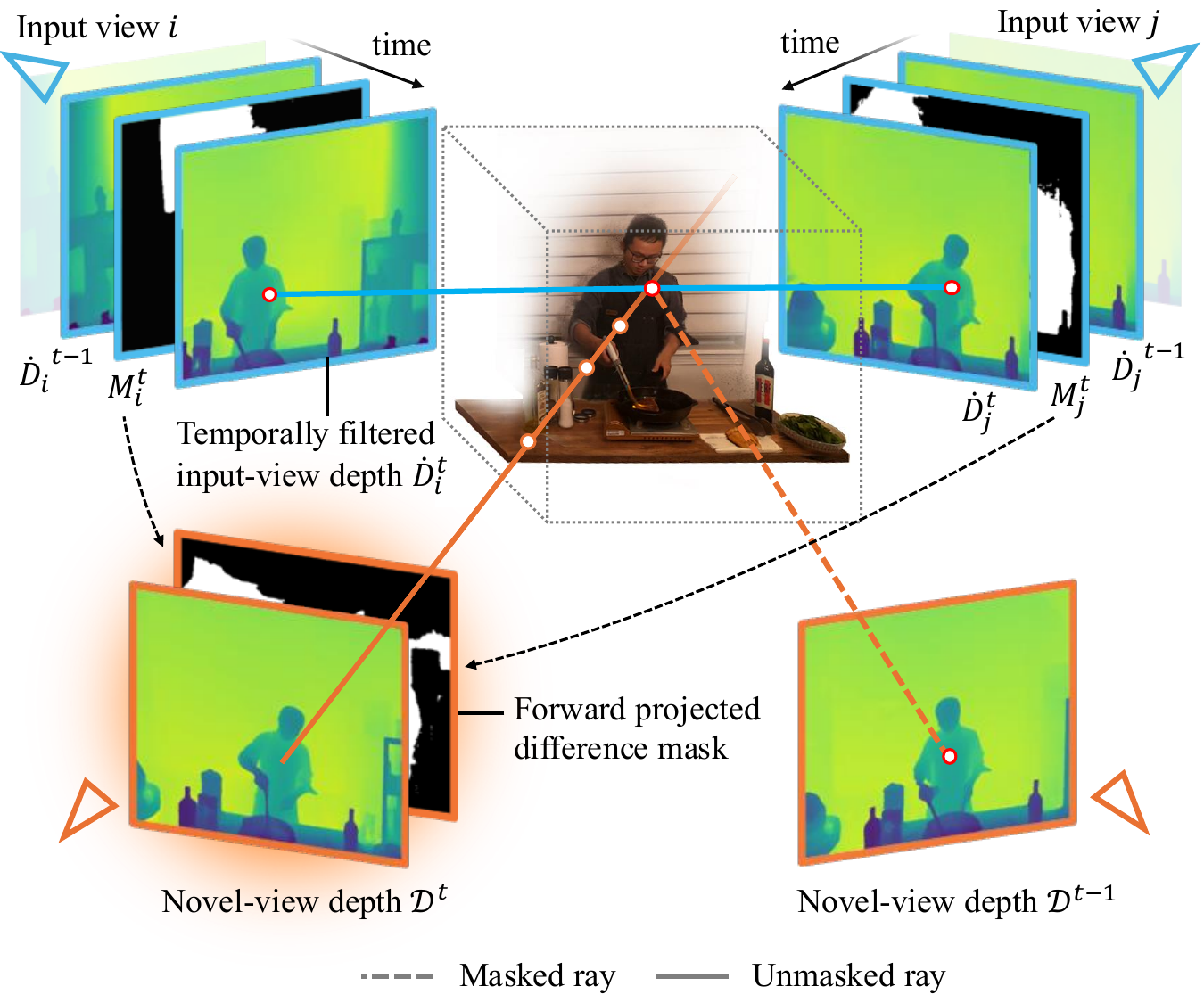}
\vspace{-3mm}
\caption{\textbf{View-temporal consistency in TSDF depth.} The global TSDF provides view consistency. Temporal consistency is encouraged by filtering the input depth at consecutive frames using color difference masks for dynamic regions. Additionally, these masks are used to integrate the rendered depth from the previous frame into the TSDF, ensuring consistency across view and time.}

\label{fig:tsdf}
\vspace{-3mm}
\end{figure}

\subsection{Multi-view Temporally Consistent Depth}
\label{subsec:tsdf}
The depth $\mathcal{D}^t$ for the novel view at time $t$ is rendered by raymarching through a TSDF that fuses the depth of the $K$ closest input views. We use an image-based TSDF \cite{lawrence2021project} that integrates depth along each ray on the fly and, thus, avoids creating an explicit 3D voxel grid. This allows it to efficiently generate high-resolution depth maps. Fusing all input depths into a global structure like a TSDF guarantees that, for a given time $t$, the rendered depth remains view-consistent. 

To encourage temporal consistency in the TSDF, we suppress depth variations across the static regions in each input depth map $D_n^t$. Specifically, given the set of $N$ input views $\{I_n^t\}$, we exploit the fact that the camera poses remain fixed across time to compute a soft difference mask between the current and previous frame of each view via
\begin{align}
M_n^t \!=\! \min\left(\frac{|I_n^t - I_n^{t-1}|}{\lambda_\text{t}} + \beta, 1\right) \text{,}
\label{eqn:difference-masks}
\end{align}
where $\beta \!=\! 0.6$ and $\lambda_\text{t} \!=\! 0.7$ are empirical hyperparameters that determine sensitivity to color changes. We compute the difference masks at quarter resolution, apply a 3$\times$3 max-pooling filter, and then upscale them back to their original size using bilinear interpolation. Given the difference masks, we compute temporally filtered depths in each view using
\begin{equation}
	\dot{D}^{t}_{n} = M^{t}_{n} \cdot D^{t}_{n} + (1-M^{t}_{n}) \cdot \dot{D}^{t-1}_{n} \text{.}
\end{equation}

While the global nature of a TSDF guarantees that the rendered depth maps remain view-consistent, in practice, however, we only fuse the depth maps of the $K$ closest input views. Consequently, minor changes in the TSDF depth can still occur as the novel viewpoint -- and, thus, the set of $K$ closest views -- changes. Increasing $K$ ameliorates these residual inconsistencies, but at increased computational cost.
Instead, we propose to use the rendered output of the previous frame $\mathcal{D}^{t-1}$ as an additional depth input to the TSDF at frame $t$. To account for dynamic regions, we forward splat the difference masks $M^{t}_{n}$ from \cref{eqn:difference-masks} into the novel view, and use the maximum mask value across the views as the TSDF blending weight for $\mathcal{D}^{t-1}$.  

Thus, the output depth map is obtained by integrating temporally filtered input-view depths, and the depth output of the previous frame in an image-based TSDF (\cref{fig:tsdf}):\looseness-1
\begin{align}
\mathcal{D}^t = \mathrm{Raymarch}(~\mathrm{TSDF}( \{\dot{D}_k^t\}, \mathcal{D}^{t-1})~).
\end{align}
This depth map, pushed to be consistent in space and time, guides the blending of the forward-rendered views $\{\mathcal{I}_k^t\}$.
For further details, please refer to our supplementary document.

\def\tccdynerffwd{0.814}
\def\tccdynerfibr{0.834}
\def\tccdynerfgps{0.771}
\def\tccdynerfenerf{0.867}
\def\tccdynerfours{0.872}

\def\tccenerffwd{0.850}
\def\tccenerfibr{0.885}
\def\tccenerfgps{0.812}
\def\tccenerfenerf{0.883}
\def\tccenerfours{0.897}

\def\tccdmvfwd{0.873}
\def\tccdmvibr{0.901}
\def\tccdmvgps{0.870}
\def\tccdmvenerf{0.892}
\def\tccdmvours{0.915}

\def\strreddynerffwd{2796}
\def\strreddynerfibr{737.7}
\def\strreddynerfgps{693.0}
\def\strreddynerfenerf{76.14}
\def\strreddynerfours{68.76}

\def\strredenerffwd{2205}
\def\strredenerfibr{65.76}
\def\strredenerfgps{299.7}
\def\strredenerfenerf{32.31}
\def\strredenerfours{44.49}

\def\strreddmvfwd{1192}
\def\strreddmvibr{341.4}
\def\strreddmvgps{637.4}
\def\strreddmvenerf{137.0}
\def\strreddmvours{15.85}

\def\sdtldynerffwd{0.214}
\def\sdtldynerfibr{0.208}
\def\sdtldynerfgps{0.298}
\def\sdtldynerfenerf{0.147}
\def\sdtldynerfours{0.139}

\def\sdtlenerffwd{0.271}
\def\sdtlenerfibr{0.193}
\def\sdtlenerfgps{0.259}
\def\sdtlenerfenerf{0.189}
\def\sdtlenerfours{0.193}

\def\sdtldmvfwd{0.433}
\def\sdtldmvibr{0.115}
\def\sdtldmvgps{1.233}
\def\sdtldmvenerf{0.144}
\def\sdtldmvours{0.105}

\def\sdvldynerffwd{5.696}
\def\sdvldynerfibr{7.233}
\def\sdvldynerfgps{2.254}
\def\sdvldynerfenerf{2.710}
\def\sdvldynerfours{1.748}

\def\sdvlenerffwd{2.560}
\def\sdvlenerfibr{2.003}
\def\sdvlenerfgps{2.159}
\def\sdvlenerfenerf{3.944}
\def\sdvlenerfours{1.970}

\def\sdvldmvfwd{0.769}
\def\sdvldmvibr{2.340}
\def\sdvldmvgps{1.830}
\def\sdvldmvenerf{0.754}
\def\sdvldmvours{0.449}

\fpmin{\tccmindynerf}{\tccdynerffwd}{\tccdynerfibr}{\tccdynerfgps}{\tccdynerfenerf}{\tccdynerfours}
\fpmax{\tccmaxdynerf}{\tccdynerffwd}{\tccdynerfibr}{\tccdynerfgps}{\tccdynerfenerf}{\tccdynerfours}
\fpnorm{\tccdynerf}{\tccmindynerf}{\tccmaxdynerf}

\fpmin{\tccminenerf}{\tccenerffwd}{\tccenerfibr}{\tccenerfgps}{\tccenerfenerf}{\tccenerfours}
\fpmax{\tccmaxenerf}{\tccenerffwd}{\tccenerfibr}{\tccenerfgps}{\tccenerfenerf}{\tccenerfours}
\fpnorm{\tccenerf}{\tccminenerf}{\tccmaxenerf}

\fpmin{\tccmindmv}{\tccdmvfwd}{\tccdmvibr}{\tccdmvgps}{\tccdmvenerf}{\tccdmvours}
\fpmax{\tccmaxdmv}{\tccdmvfwd}{\tccdmvibr}{\tccdmvgps}{\tccdmvenerf}{\tccdmvours}
\fpnorm{\tccdmv}{\tccmindmv}{\tccmaxdmv}

\fpmin{\strredmindynerf}{\strreddynerffwd}{\strreddynerfibr}{\strreddynerfgps}{\strreddynerfenerf}{\strreddynerfours}
\fpmax{\strredmaxdynerf}{\strreddynerffwd}{\strreddynerfibr}{\strreddynerfgps}{\strreddynerfenerf}{\strreddynerfours}
\fpnorminv{\strreddynerf}{\strredmindynerf}{\strredmaxdynerf}

\fpmin{\strredminenerf}{\strredenerffwd}{\strredenerfibr}{\strredenerfgps}{\strredenerfenerf}{\strredenerfours}
\fpmax{\strredmaxenerf}{\strredenerffwd}{\strredenerfibr}{\strredenerfgps}{\strredenerfenerf}{\strredenerfours}
\fpnorminv{\strredenerf}{\strredminenerf}{\strredmaxenerf}

\fpmin{\strredmindmv}{\strreddmvfwd}{\strreddmvibr}{\strreddmvgps}{\strreddmvenerf}{\strreddmvours}
\fpmax{\strredmaxdmv}{\strreddmvfwd}{\strreddmvibr}{\strreddmvgps}{\strreddmvenerf}{\strreddmvours}
\fpnorminv{\strreddmv}{\strredmindmv}{\strredmaxdmv}

\fpmin{\sdtlmindynerf}{\sdtldynerffwd}{\sdtldynerfibr}{\sdtldynerfgps}{\sdtldynerfenerf}{\sdtldynerfours}
\fpmax{\sdtlmaxdynerf}{\sdtldynerffwd}{\sdtldynerfibr}{\sdtldynerfgps}{\sdtldynerfenerf}{\sdtldynerfours}
\fpnorminv{\sdtldynerf}{\sdtlmindynerf}{\sdtlmaxdynerf}

\fpmin{\sdtlminenerf}{\sdtlenerffwd}{\sdtlenerfibr}{\sdtlenerfgps}{\sdtlenerfenerf}{\sdtlenerfours}
\fpmax{\sdtlmaxenerf}{\sdtlenerffwd}{\sdtlenerfibr}{\sdtlenerfgps}{\sdtlenerfenerf}{\sdtlenerfours}
\fpnorminv{\sdtlenerf}{\sdtlminenerf}{\sdtlmaxenerf}

\fpmin{\sdtlmindmv}{\sdtldmvfwd}{\sdtldmvibr}{\sdtldmvgps}{\sdtldmvenerf}{\sdtldmvours}
\fpmax{\sdtlmaxdmv}{\sdtldmvfwd}{\sdtldmvibr}{\sdtldmvgps}{\sdtldmvenerf}{\sdtldmvours}
\fpnorminv{\sdtldmv}{\sdtlmindmv}{\sdtlmaxdmv}

\fpmin{\sdvlmindynerf}{\sdvldynerffwd}{\sdvldynerfibr}{\sdvldynerfgps}{\sdvldynerfenerf}{\sdvldynerfours}
\fpmax{\sdvlmaxdynerf}{\sdvldynerffwd}{\sdvldynerfibr}{\sdvldynerfgps}{\sdvldynerfenerf}{\sdvldynerfours}
\fpnorminv{\sdvldynerf}{\sdvlmindynerf}{\sdvlmaxdynerf}

\fpmin{\sdvlminenerf}{\sdvlenerffwd}{\sdvlenerfibr}{\sdvlenerfgps}{\sdvlenerfenerf}{\sdvlenerfours}
\fpmax{\sdvlmaxenerf}{\sdvlenerffwd}{\sdvlenerfibr}{\sdvlenerfgps}{\sdvlenerfenerf}{\sdvlenerfours}
\fpnorminv{\sdvlenerf}{\sdvlminenerf}{\sdvlmaxenerf}

\fpmin{\sdvlmindmv}{\sdvldmvfwd}{\sdvldmvibr}{\sdvldmvgps}{\sdvldmvenerf}{\sdvldmvours}
\fpmax{\sdvlmaxdmv}{\sdvldmvfwd}{\sdvldmvibr}{\sdvldmvgps}{\sdvldmvenerf}{\sdvldmvours}
\fpnorminv{\sdvldmv}{\sdvlmindmv}{\sdvlmaxdmv}

	\setlength{\tabcolsep}{1.8mm}
	\begin{table*}
		\caption{\label{table:quantitative-vt-results}%
			\textbf{Quantitative comparison of view and temporal consistency.}
			We highlight the 50\% best metrics in blue, proportional to their percentile.
			Our method achieves the best metrics on DyNeRF and D3DMV, and partially on the ENeRF-Outdoor dataset.
			Please refer to the qualitative results on view and temporal consistency in \cref{fig:results-view-consistency,fig:results-temporal-consistency}, and to the supplementary material for video results.
			}
		\centering\small
		\begin{tabular}{lcccccccccccccc}
			\toprule
			& \multicolumn{4}{c}{\textbf{DyNeRF \cite{li2022neural}}} &
			& \multicolumn{4}{c}{\textbf{ENeRF-Outdoor \cite{lin2022efficient}}} &
			& \multicolumn{4}{c}{\textbf{D3DMV \cite{lin2021deep}}} \\
			\midrule
			Method & TCC$\uparrow$ & STED$\downarrow$ & SDT$\downarrow$ &SDV$\downarrow$ & & TCC$\uparrow$ & STED$\downarrow$ & SDT$\downarrow$ &SDV$\downarrow$ & & TCC$\uparrow$ & STED$\downarrow$ & SDT$\downarrow$ &SDV$\downarrow$ \\
			\midrule
			FWD~\cite{cao2022fwd} & \hspace{-2mm}\tccdynerf{\tccdynerffwd} & \hspace{-2mm}\strreddynerf{\strreddynerffwd}\phantom{.00} & \hspace{-2mm}\sdtldynerf{\sdtldynerffwd} & \hspace{-2mm}\sdvldynerf{\sdvldynerffwd} & & \hspace{-2mm}\tccenerf{\tccenerffwd} & \hspace{-2mm}\strredenerf{\strredenerffwd}\phantom{.00} & \hspace{-2mm}\sdtlenerf{\sdtlenerffwd} & \hspace{-2mm}\sdvlenerf{\sdvlenerffwd} & & \hspace{-2mm}\tccdmv{\tccdmvfwd} & \hspace{-2mm}\strreddmv{\strreddmvfwd}\phantom{.00} & \hspace{-2mm}\sdtldmv{\sdtldmvfwd} & \hspace{-2mm}\sdvldmv{\sdvldmvfwd} \\

			IBRNet~\cite{wang2021ibrnet} & \hspace{-2mm}\tccdynerf{\tccdynerfibr} & \hspace{-2mm}\pad\strreddynerf{\strreddynerfibr}\pad & \hspace{-2mm}\sdtldynerf{\sdtldynerfibr} & \hspace{-2mm}\sdvldynerf{\sdvldynerfibr} & & \hspace{-2mm}\tccenerf{\tccenerfibr} & \hspace{-2mm}\pad\pad\strredenerf{\strredenerfibr} & \hspace{-2mm}\sdtlenerf{\sdtlenerfibr} & \hspace{-2mm}\sdvlenerf{\sdvlenerfibr} & & \hspace{-2mm}\tccdmv{\tccdmvibr} & \hspace{-2mm}\pad\strreddmv{\strreddmvibr}\pad & \hspace{-2mm}\sdtldmv{\sdtldmvibr} & \hspace{-2mm}\sdvldmv{\sdvldmvibr} \\

			GPSG~\cite{zheng2024gps} & \hspace{-2mm}\tccdynerf{\tccdynerfgps} & \hspace{-2mm}\pad\strreddynerf{\strreddynerfgps}\pad & \hspace{-2mm}\sdtldynerf{\sdtldynerfgps} & \hspace{-2mm}\sdvldynerf{\sdvldynerfgps} & & \hspace{-2mm}\tccenerf{\tccenerfgps} & \hspace{-2mm}\pad\strredenerf{\strredenerfgps}\pad & \hspace{-2mm}\sdtlenerf{\sdtlenerfgps} & \hspace{-2mm}\sdvlenerf{\sdvlenerfgps} & & \hspace{-2mm}\tccdmv{\tccdmvgps} & \hspace{-2mm}\pad\strreddmv{\strreddmvgps}\pad & \hspace{-2mm}\sdtldmv{\sdtldmvgps} & \hspace{-2mm}\sdvldmv{\sdvldmvgps} \\

			ENeRF~\cite{lin2022efficient} & \hspace{-2mm}\tccdynerf{\tccdynerfenerf} & \hspace{-2mm}\pad\pad\strreddynerf{\strreddynerfenerf} & \hspace{-2mm}\sdtldynerf{\sdtldynerfenerf} & \hspace{-2mm}\sdvldynerf{\sdvldynerfenerf} & & \hspace{-2mm}\tccenerf{\tccenerfenerf} & \hspace{-2mm}\pad\pad\strredenerf{\strredenerfenerf} & \hspace{-2mm}\sdtlenerf{\sdtlenerfenerf} & \hspace{-2mm}\sdvlenerf{\sdvlenerfenerf} & & \hspace{-2mm}\tccdmv{\tccdmvenerf} & \hspace{-2mm}\pad\strreddmv{\strreddmvenerf}\pad & \hspace{-2mm}\sdtldmv{\sdtldmvenerf} & \hspace{-2mm}\sdvldmv{\sdvldmvenerf} \\

			Ours & \hspace{-2mm}\tccdynerf{\tccdynerfours} & \hspace{-2mm}\pad\pad\strreddynerf{\strreddynerfours} & \hspace{-2mm}\sdtldynerf{\sdtldynerfours} & \hspace{-2mm}\sdvldynerf{\sdvldynerfours} & & \hspace{-2mm}\tccenerf{\tccenerfours} & \hspace{-2mm}\pad\pad\strredenerf{\strredenerfours} & \hspace{-2mm}\sdtlenerf{\sdtlenerfours} & \hspace{-2mm}\sdvlenerf{\sdvlenerfours} & & \hspace{-2mm}\tccdmv{\tccdmvours} & \hspace{-2mm}\pad\pad\strreddmv{\strreddmvours} & \hspace{-2mm}\sdtldmv{\sdtldmvours} & \hspace{-2mm}\sdvldmv{\sdvldmvours}\\
			\bottomrule
		\end{tabular}
	\end{table*}

\subsection{Geometry-guided Blending Network}
\label{subsec:blending}
The final novel-view output of our method, $\mathcal{I}^t$, is generated by a weighted blending of the $K$ forward-rendered images $\{\mathcal{I}_k^t \}$. We use per-pixel weights $\{w_k^t \in \mathbb{R}^{H\times W}\}$. In addition, to deal with disoccluded regions in the novel view, we further blend the result with a background image $\mathcal{I}_{\text{BG}}^t$ using blending weights $w_{\text{BG}}^t \in \mathbb{R}^{H\times W}$. Thus,
\begin{align}
\mathcal{I}_{\text{FG}}^t &= \sum_{k=1}^{K}
\frac{\mathcal{I}_k^t \cdot \alpha_k^t \cdot w_k^t}{\alpha_k^t \cdot w_k^t} \quad \text{and}
\label{eq:fg_blending} \\
\mathcal{I}^t &=  (1 - w_{\text{BG}}^t) \cdot \mathcal{I}_{\text{FG}}^t + w_{\text{BG}}^t \cdot \mathcal{I}_{\text{BG}}^t \text{.}
\label{eq:bg_blending}
\end{align}

We use a neural network $\Theta(\cdot)$ to predict the blending weights $\{w_k^t\}$ and $w_{\text{BG}}^t$,  and the background image $\mathcal{I}_{\text{BG}}^t$.
The prediction is guided by the consistent TSDF depth $\mathcal{D}^t$.
To achieve this, we provide $\mathcal{D}^t$, and the forward-rendered depth maps $\{\mathcal{D}_k^t\}$ as inputs to the network $\Theta(\cdot)$.
To encourage the blending weights to reconstruct $\mathcal{D}^t$, we use the loss
\begin{align}
\mathrm{L}_\text{Depth} = \frac{1}{HWK}~\Bigl\lVert\mathcal{D}^t - \sum_{k=1}^{K}
\frac{\mathcal{D}_k^t \cdot \alpha_k^t \cdot w_k^t}{\alpha_k^t \cdot w_k^t} \Bigr\rVert_1 \text{.}
\label{eq:depth_blending}
\end{align}
We use this loss in conjunction with image reconstruction losses, as described below.
Thus, $\mathrm{L}_\text{Depth}$ provides a soft regularization that encourages the network to predict the unknown novel-view image using the known TSDF depth as a target.
Therefore, by ensuring the TSDF depth map $\mathcal{D}^t$ remains view-temporally consistent, the reconstructed image $\mathcal{I}^t$ is guided to be consistent, too.

The network $\Theta(\cdot)$ is a four-layer U-Net \cite{ronneberger2015u}. The input to the network consist of the forward-rendered images $\{\mathcal{I}_k^t\}$, depth maps $\{D_k^t\}$ and alpha maps $\{\mathcal{\alpha
}_k^t\}$, and the TSDF depth $\mathcal{D}^t$. Additionally, we provide camera orientation information in the form of camera distances, and dot products between the viewing angles and normals. Please refer to the supplement for details about the network input and architecture.

The total training loss function is
\begin{align}
\mathrm{L} = 0.8 \!\cdot\! \mathrm{L}_{1} + 0.2 \!\cdot\! \mathrm{L}_\text{SSIM} + 0.1 \!\cdot\! \mathrm{L}_\text{Depth} + 0.1 \!\cdot\! \mathrm{L}_\text{Mask} \text{,}
\end{align}
where $\mathrm{L}_1$ and $\mathrm{L}_\text{SSIM}$ are the L1 and SSIM \cite{wang2004image} losses on the reconstructed image $\mathcal{I}^t$, and $\mathrm{L}_{\text{Mask}}$ encourages the background weights $w_\text{BG}^t$ to be higher in disoccluded regions. It is computed as the mean absolute difference between  $w_\text{BG}^t$, and a binary mask of the empty regions in $\mathcal{I}^t$.
\section{Evaluation}
\label{sec:experiments}

\begin{figure*}[t]
\vspace{0em}
  \centering
  \includegraphics[trim=0cm 6.5cm 11.7cm 0cm, clip=true, width=\linewidth]{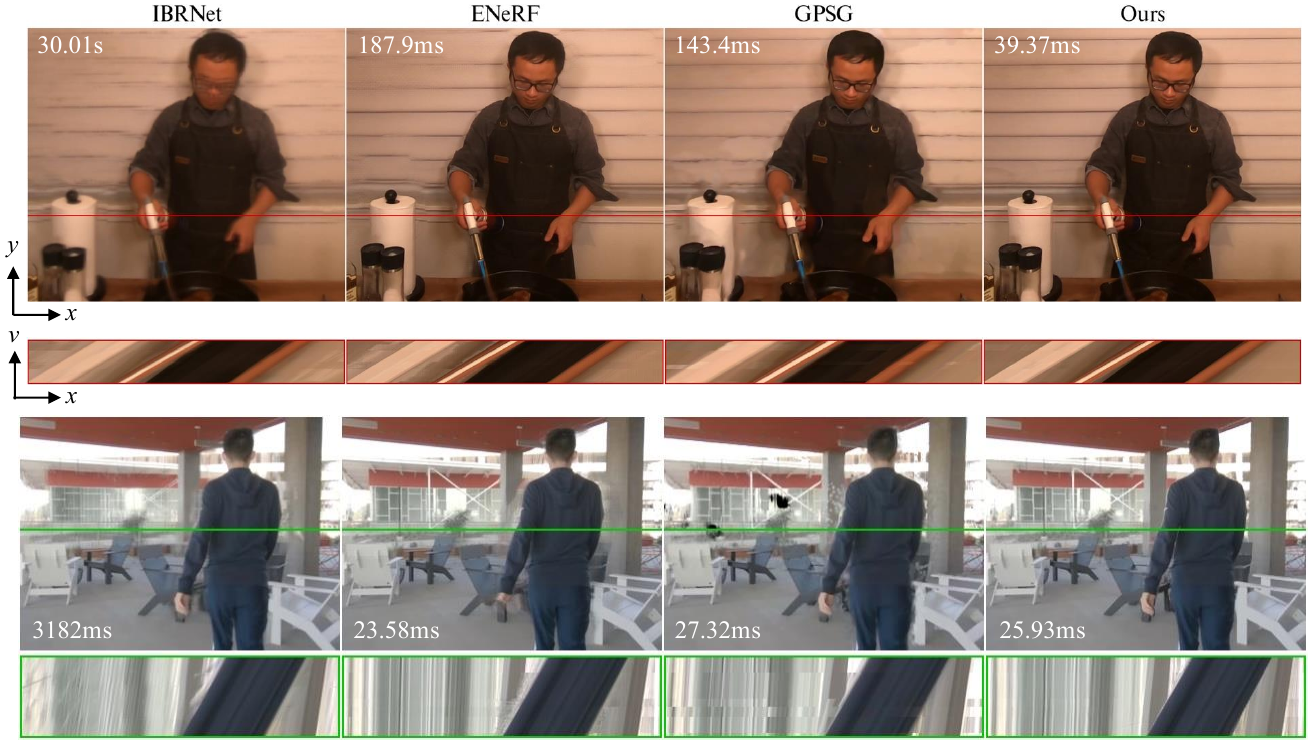}  
  \caption{\textbf{Qualitative comparison of view consistency.} We compare view consistency using epipolar plane images (EPI). We generate the EPI for each method by rendering novel views along a horizontally-translating camera path (the $v$ dimension). For a fixed image row $y$ an EPI then represents a slice of the scene in the space-view dimensions $x$–$v$, allowing a subset of points to be visualized across views as sloping lines. All baseline methods show sudden changes along EPI lines indicating view inconsistency. Our method generates smooth and continuous EPI lines, showing that it maintains view consistency as the camera translates. }
\label{fig:results-view-consistency}%
\vspace{-1em}
\end{figure*}

\begin{figure*}[t]
\vspace{1em}
\includegraphics[trim=0cm 12.0cm 11.7cm 0cm, clip=true, width=1.0\textwidth]{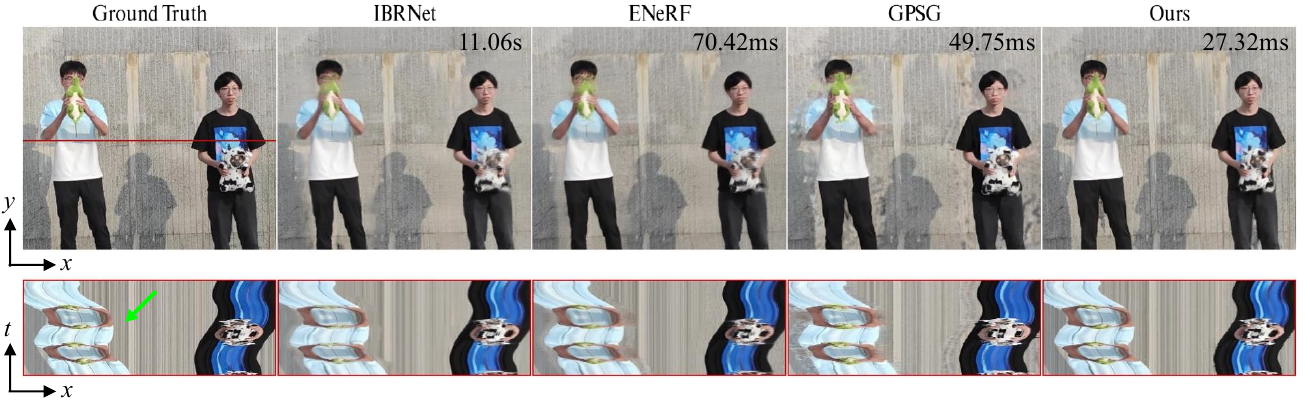}\vspace{-1em}
\caption{\label{fig:results-temporal-consistency}%
  \textbf{Qualitative comparison of temporal consistency.} We compare temporal consistency using a slice of the scene in the space-time dimensions $x$–$t$. This is generated by rendering novel views at a fixed viewpoint while time moves forward. The baseline methods either suffer from noise in the temporal slices indicating temporal inconsistency (ENeRF, GPSG), or show blurred results due to low \emph{spatial} reconstruction quality (IBRNet). Our method maintains temporal consistency while preserving high frequency spatial details. Please refer to the supplementary material for video results on all datasets.
}
\vspace{-1em}
\end{figure*}

\def\psnrdynerffwd{16.97}
\def\psnrdynerfibr{22.49}
\def\psnrdynerfgps{25.46}
\def\psnrdynerfenerf{29.51}
\def\psnrdynerfours{31.17}

\def\psnrenerffwd{17.01}
\def\psnrenerfibr{22.96}
\def\psnrenerfgps{23.10}
\def\psnrenerfenerf{25.79}
\def\psnrenerfours{24.01}

\def\psnrdmvfwd{19.80}
\def\psnrdmvibr{20.71}
\def\psnrdmvgps{23.77}
\def\psnrdmvenerf{27.79}
\def\psnrdmvours{32.58}

\def\ssimdynerffwd{0.667}
\def\ssimdynerfibr{0.826}
\def\ssimdynerfgps{0.866}
\def\ssimdynerfenerf{0.927}
\def\ssimdynerfours{0.936}

\def\ssimenerffwd{0.395}
\def\ssimenerfibr{0.663}
\def\ssimenerfgps{0.687}
\def\ssimenerfenerf{0.722}
\def\ssimenerfours{0.703}

\def\ssimdmvfwd{0.585}
\def\ssimdmvibr{0.710}
\def\ssimdmvgps{0.884}
\def\ssimdmvenerf{0.879}
\def\ssimdmvours{0.936}

\def\lpipsdynerffwd{0.554}
\def\lpipsdynerfibr{0.347}
\def\lpipsdynerfgps{0.309}
\def\lpipsdynerfenerf{0.196}
\def\lpipsdynerfours{0.191}

\def\lpipsenerffwd{0.699}
\def\lpipsenerfibr{0.224}
\def\lpipsenerfgps{0.231}
\def\lpipsenerfenerf{0.167}
\def\lpipsenerfours{0.155}

\def\lpipsdmvfwd{0.640}
\def\lpipsdmvibr{0.358}
\def\lpipsdmvgps{0.156}
\def\lpipsdmvenerf{0.157}
\def\lpipsdmvours{0.098}

\def\ldynerffwd{25.76}
\def\ldynerfibr{11.83}
\def\ldynerfgps{8.00}
\def\ldynerfenerf{6.06}
\def\ldynerfours{5.09}

\def\lenerffwd{24.39}
\def\lenerfibr{13.06}
\def\lenerfgps{10.96}
\def\lenerfenerf{9.84}
\def\lenerfours{12.02}

\def\ldmvfwd{19.39}
\def\ldmvibr{11.98}
\def\ldmvgps{7.70}
\def\ldmvenerf{6.01}
\def\ldmvours{3.60}

\fpmin{\psnrmindynerf}{\psnrdynerffwd}{\psnrdynerfibr}{\psnrdynerfgps}{\psnrdynerfenerf}{\psnrdynerfours}
\fpmax{\psnrmaxdynerf}{\psnrdynerffwd}{\psnrdynerfibr}{\psnrdynerfgps}{\psnrdynerfenerf}{\psnrdynerfours}
\fpnorm{\psnrdynerf}{\psnrmindynerf}{\psnrmaxdynerf}

\fpmin{\psnrminenerf}{\psnrenerffwd}{\psnrenerfibr}{\psnrenerfgps}{\psnrenerfenerf}{\psnrenerfours}
\fpmax{\psnrmaxenerf}{\psnrenerffwd}{\psnrenerfibr}{\psnrenerfgps}{\psnrenerfenerf}{\psnrenerfours}
\fpnorm{\psnrenerf}{\psnrminenerf}{\psnrmaxenerf}

\fpmin{\psnrmindmv}{\psnrdmvfwd}{\psnrdmvibr}{\psnrdmvgps}{\psnrdmvenerf}{\psnrdmvours}
\fpmax{\psnrmaxdmv}{\psnrdmvfwd}{\psnrdmvibr}{\psnrdmvgps}{\psnrdmvenerf}{\psnrdmvours}
\fpnorm{\psnrdmv}{\psnrmindmv}{\psnrmaxdmv}

\fpmin{\ssimmindynerf}{\ssimdynerffwd}{\ssimdynerfibr}{\ssimdynerfgps}{\ssimdynerfenerf}{\ssimdynerfours}
\fpmax{\ssimmaxdynerf}{\ssimdynerffwd}{\ssimdynerfibr}{\ssimdynerfgps}{\ssimdynerfenerf}{\ssimdynerfours}
\fpnorm{\ssimdynerf}{\ssimmindynerf}{\ssimmaxdynerf}

\fpmin{\ssimminenerf}{\ssimenerffwd}{\ssimenerfibr}{\ssimenerfgps}{\ssimenerfenerf}{\ssimenerfours}
\fpmax{\ssimmaxenerf}{\ssimenerffwd}{\ssimenerfibr}{\ssimenerfgps}{\ssimenerfenerf}{\ssimenerfours}
\fpnorm{\ssimenerf}{\ssimminenerf}{\ssimmaxenerf}

\fpmin{\ssimmindmv}{\ssimdmvfwd}{\ssimdmvibr}{\ssimdmvgps}{\ssimdmvenerf}{\ssimdmvours}
\fpmax{\ssimmaxdmv}{\ssimdmvfwd}{\ssimdmvibr}{\ssimdmvgps}{\ssimdmvenerf}{\ssimdmvours}
\fpnorm{\ssimdmv}{\ssimmindmv}{\ssimmaxdmv}

\fpmin{\lpipsmindynerf}{\lpipsdynerffwd}{\lpipsdynerfibr}{\lpipsdynerfgps}{\lpipsdynerfenerf}{\lpipsdynerfours}
\fpmax{\lpipsmaxdynerf}{\lpipsdynerffwd}{\lpipsdynerfibr}{\lpipsdynerfgps}{\lpipsdynerfenerf}{\lpipsdynerfours}
\fpnorminv{\lpipsdynerf}{\lpipsmindynerf}{\lpipsmaxdynerf}

\fpmin{\lpipsminenerf}{\lpipsenerffwd}{\lpipsenerfibr}{\lpipsenerfgps}{\lpipsenerfenerf}{\lpipsenerfours}
\fpmax{\lpipsmaxenerf}{\lpipsenerffwd}{\lpipsenerfibr}{\lpipsenerfgps}{\lpipsenerfenerf}{\lpipsenerfours}
\fpnorminv{\lpipsenerf}{\lpipsminenerf}{\lpipsmaxenerf}

\fpmin{\lpipsmindmv}{\lpipsdmvfwd}{\lpipsdmvibr}{\lpipsdmvgps}{\lpipsdmvenerf}{\lpipsdmvours}
\fpmax{\lpipsmaxdmv}{\lpipsdmvfwd}{\lpipsdmvibr}{\lpipsdmvgps}{\lpipsdmvenerf}{\lpipsdmvours}
\fpnorminv{\lpipsdmv}{\lpipsmindmv}{\lpipsmaxdmv}

\fpmin{\lmindynerf}{\ldynerffwd}{\ldynerfibr}{\ldynerfgps}{\ldynerfenerf}{\ldynerfours}
\fpmax{\lmaxdynerf}{\ldynerffwd}{\ldynerfibr}{\ldynerfgps}{\ldynerfenerf}{\ldynerfours}
\fpnorminv{\ldynerf}{\lmindynerf}{\lmaxdynerf}

\fpmin{\lminenerf}{\lenerffwd}{\lenerfibr}{\lenerfgps}{\lenerfenerf}{\lenerfours}
\fpmax{\lmaxenerf}{\lenerffwd}{\lenerfibr}{\lenerfgps}{\lenerfenerf}{\lenerfours}
\fpnorminv{\lenerf}{\lminenerf}{\lmaxenerf}

\fpmin{\lmindmv}{\ldmvfwd}{\ldmvibr}{\ldmvgps}{\ldmvenerf}{\ldmvours}
\fpmax{\lmaxdmv}{\ldmvfwd}{\ldmvibr}{\ldmvgps}{\ldmvenerf}{\ldmvours}
\fpnorminv{\ldmv}{\lmindmv}{\lmaxdmv}

\setlength{\tabcolsep}{1.2mm}
\begin{table*}
\vspace{6mm}
\caption{\label{table:quantitative-results}%
	\textbf{Quantitative comparison of novel-view synthesis on the DyNeRF, ENeRF-Outdoor, and D3DMV datasets.}
	We highlight the 50\% best metrics in blue, proportional to their percentile.
	Our method achieves the best metrics for the DyNeRF and D3DMV datasets.
	Please refer to the qualitative results on novel-view synthesis in \cref{fig:results-qualitative}.
}
\centering\small
\begin{tabular}{lcccccccccccccc}
\toprule
& \multicolumn{4}{c}{\textbf{DyNeRF \cite{li2022neural}}} &
& \multicolumn{4}{c}{\textbf{ENeRF-Outdoor \cite{lin2022efficient}}} &
& \multicolumn{4}{c}{\textbf{D3DMV \cite{lin2021deep}}} \\
\midrule
Method & \hspace{1mm}PSNR$\uparrow$ & \hspace{1mm}SSIM$\uparrow$ & \hspace{1mm}LPIPS$\downarrow$ & \hspace{1mm}L1$\downarrow$ & & \hspace{1mm}PSNR$\uparrow$ & \hspace{1mm}SSIM$\uparrow$ & \hspace{1mm}LPIPS$\downarrow$ & \hspace{1mm}L1$\downarrow$ & & \hspace{1mm}PSNR$\uparrow$ & \hspace{1mm}SSIM$\uparrow$ & \hspace{1mm}LPIPS$\downarrow$ & \hspace{1mm}L1$\downarrow$ \\
\midrule

FWD~\cite{cao2022fwd} & \hspace{-2mm}\psnrdynerf{\psnrdynerffwd} & \hspace{-2mm}\ssimdynerf{\ssimdynerffwd} & \hspace{-2mm}\lpipsdynerf{\lpipsdynerffwd} & \hspace{-2mm}\ldynerf{\ldynerffwd} & & \hspace{-2mm}\psnrenerf{\psnrenerffwd} & \hspace{-2mm}\ssimenerf{\ssimenerffwd} & \hspace{-2mm}\lpipsenerf{\lpipsenerffwd} & \hspace{-2mm}\lenerf{\lenerffwd} & & \hspace{-2mm}\psnrdmv{\psnrdmvfwd} & \hspace{-2mm}\ssimdmv{\ssimdmvfwd} & \hspace{-2mm}\lpipsdmv{\lpipsdmvfwd} & \hspace{-2mm}\ldmv{\ldmvfwd} \\

IBRNet~\cite{wang2021ibrnet} & \hspace{-2mm}\psnrdynerf{\psnrdynerfibr} & \hspace{-2mm}\ssimdynerf{\ssimdynerfibr} & \hspace{-2mm}\lpipsdynerf{\lpipsdynerfibr} & \hspace{-2mm}\ldynerf{\ldynerfibr} & & \hspace{-2mm}\psnrenerf{\psnrenerfibr} & \hspace{-2mm}\ssimenerf{\ssimenerfibr} & \hspace{-2mm}\lpipsenerf{\lpipsenerfibr} & \hspace{-2mm}\lenerf{\lenerfibr} & & \hspace{-2mm}\psnrdmv{\psnrdmvibr} & \hspace{-2mm}\ssimdmv{\ssimdmvibr} & \hspace{-2mm}\lpipsdmv{\lpipsdmvibr} & \hspace{-2mm}\ldmv{\ldmvibr} \\

GPSG~\cite{zheng2024gps} & \hspace{-2mm}\psnrdynerf{\psnrdynerfgps} & \hspace{-2mm}\ssimdynerf{\ssimdynerfgps} & \hspace{-2mm}\lpipsdynerf{\lpipsdynerfgps} & \hspace{-2mm}\pad\ldynerf{\ldynerfgps} & & \hspace{-2mm}\psnrenerf{\psnrenerfgps} & \hspace{-2mm}\ssimenerf{\ssimenerfgps} & \hspace{-2mm}\lpipsenerf{\lpipsenerfgps} & \hspace{-2mm}\lenerf{\lenerfgps} & & \hspace{-2mm}\psnrdmv{\psnrdmvgps} & \hspace{-2mm}\ssimdmv{\ssimdmvgps} & \hspace{-2mm}\lpipsdmv{\lpipsdmvgps} & \hspace{-2mm}\pad\ldmv{\ldmvgps} \\

ENeRF~\cite{lin2022efficient} & \hspace{-2mm}\psnrdynerf{\psnrdynerfenerf} & \hspace{-2mm}\ssimdynerf{\ssimdynerfenerf} & \hspace{-2mm}\lpipsdynerf{\lpipsdynerfenerf} & \hspace{-2mm}\pad\ldynerf{\ldynerfenerf} & & \hspace{-2mm}\psnrenerf{\psnrenerfenerf} & \hspace{-2mm}\ssimenerf{\ssimenerfenerf} & \hspace{-2mm}\lpipsenerf{\lpipsenerfenerf} & \hspace{-2mm}\pad\lenerf{\lenerfenerf} & & \hspace{-2mm}\psnrdmv{\psnrdmvenerf} & \hspace{-2mm}\ssimdmv{\ssimdmvenerf} & \hspace{-2mm}\lpipsdmv{\lpipsdmvenerf} & \hspace{-2mm}\pad\ldmv{\ldmvenerf} \\

Ours & \hspace{-2mm}\psnrdynerf{\psnrdynerfours} & \hspace{-2mm}\ssimdynerf{\ssimdynerfours} & \hspace{-2mm}\lpipsdynerf{\lpipsdynerfours} & \hspace{-2mm}\pad\ldynerf{\ldynerfours} & & \hspace{-2mm}\psnrenerf{\psnrenerfours} & \hspace{-2mm}\ssimenerf{\ssimenerfours} & \hspace{-2mm}\lpipsenerf{\lpipsenerfours} & \hspace{-2mm}\lenerf{\lenerfours} & & \hspace{-2mm}\psnrdmv{\psnrdmvours} & \hspace{-2mm}\ssimdmv{\ssimdmvours} & \hspace{-2mm}\lpipsdmv{\lpipsdmvours} & \hspace{-2mm}\pad\ldmv{\ldmvours}\\
\bottomrule
\end{tabular}
\vspace{1em}
\end{table*}

\begin{figure*}[t]
\centering%
\small
\begin{tabular}{@{}p{0.2\textwidth}@{}p{0.2\textwidth}@{}p{0.2\textwidth}@{}p{0.2\textwidth}@{}p{0.2\textwidth}@{}}
  \centering Ground Truth &
  \centering IBRNet \cite{wang2021ibrnet} &
  \centering ENeRF \cite{lin2022efficient} &
  \centering GPSG \cite{zheng2024gps} &
  \centering Ours
\end{tabular}
\includegraphics[trim=0cm 0.5cm 0cm 3cm, clip=true, width=1.0\textwidth]{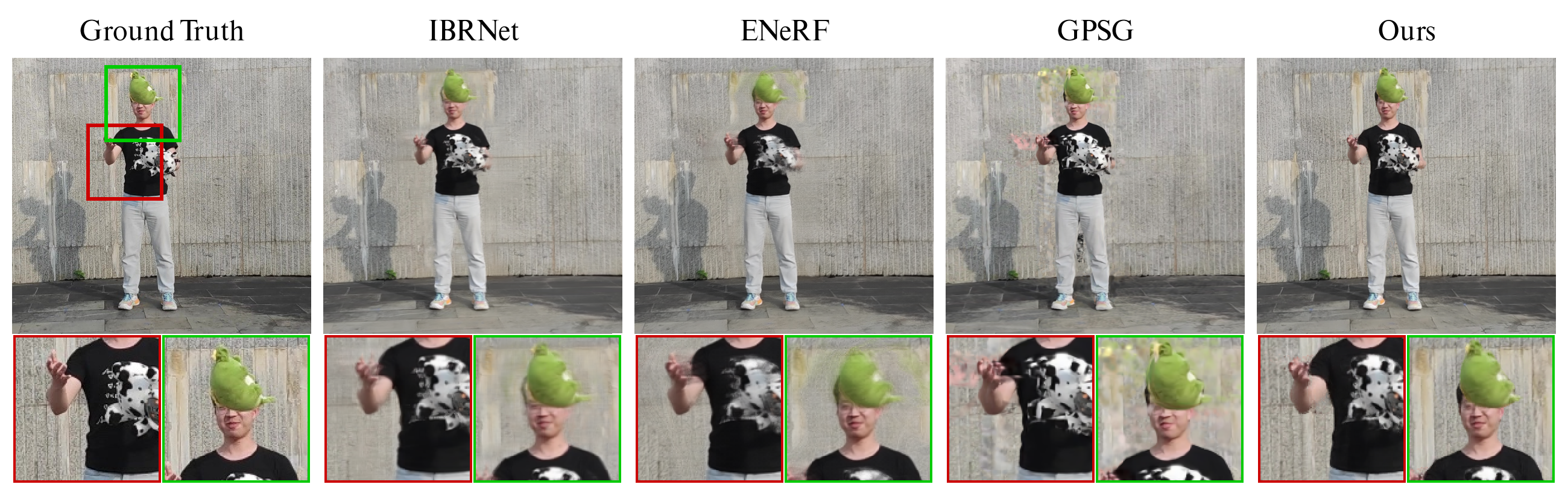}\\
\vspace{0.5em}
\label{fig:results-combined-enerf}%
\includegraphics[trim=0cm 1.0cm 0cm 0cm, clip=true, width=0.99\textwidth]{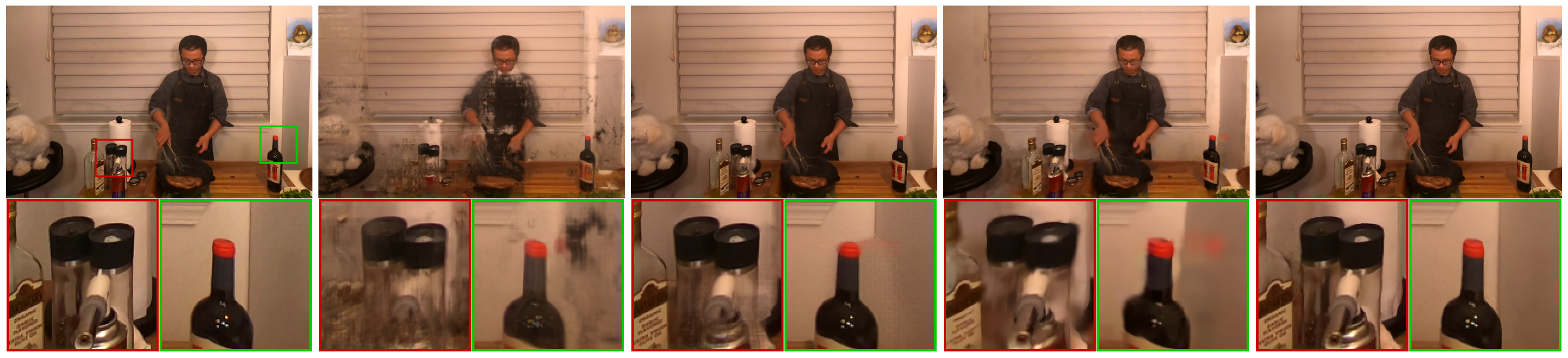}\\
\vspace{0.5em}
\label{fig:results-combined-dynerf}%
\includegraphics[trim=0cm 0.0cm 0cm 0cm, clip=true, width=0.995\textwidth]{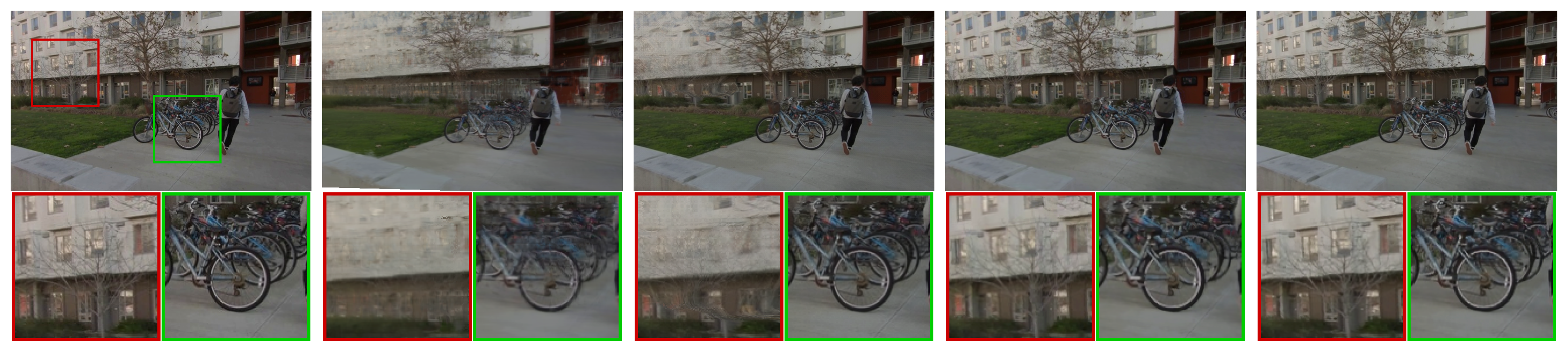}
\label{fig:results-combined-d3dmv}%
\caption{\textbf{Qualitative comparison on the ENeRF-Outdoor \cite{lin2022efficient}, DyNeRF \cite{li2022neural}, and D3DMV \cite{lin2021deep} datasets.}
The baseline methods either generate blurry results (IBRNet, ENeRF), bleed foreground colors onto the background (ENeRF, GPSG), distort edges (GPSG), or fail to reconstruct distant background objects (IBRNet, ENeRF). Our method reconstructs fine details in the background while generating sharper results for the near-range objects such as the moving humans.}
\label{fig:results-qualitative}%
\end{figure*}

\label{sec:implementation}

We implement our blending network in PyTorch, and train on an NVIDIA Tesla V100 GPU.
We use the Adam optimizer \cite{loshchilov2017decoupled} (learning rate: $2\!\times\!10^{-4}$, weight decay: $10^{-5}$) to train the network for 150K iterations in batches of size 12 on 640$\times$352 crops of ScanNet \cite{dai2017scannet}, Google Spaces \cite{flynn2019deepview}, and a training subset of the DyNeRF \cite{li2022neural} dataset.
We utilize Torch-TensorRT \cite{torchtensorrt2024} to speed up inference.
For Google Spaces and DyNeRF, which have fixed camera configurations, we compute the depth for each view using RAFT-Stereo \cite{lipson2021raft} on manually selected image pairs.
The ScanNet dataset provides depth from a sensor.
Each training iteration, the rendering target, and $K=4$ input views are randomly selected.
Our image-based TSDF is implemented in CUDA, and we use the original 3D Gaussian splatting code \cite{kerbl20233d} for forward rendering input views.

\subsection{Baselines}
\label{sec:baselines}
We compare our results to four baseline methods for online view synthesis from multi-camera video:
FWD \cite{cao2022fwd}, IBRNet \cite{wang2021ibrnet}, ENeRF \cite{lin2022efficient}, and GPSG \cite{zheng2024gps}.
We evaluate IBRNet, and ENeRF using the pretrained models provided by their authors.
The pretrained models for GPSG and FWD, however, fail to generalize to our testing datasets.
Therefore, for a fair comparison, we retrain their networks with the same training data and strategy as ours.
\subsection{Testing Datasets}
\label{sec:testing-datasets}

We evaluate all methods on three datasets: DyNeRF \cite{li2022neural} (1352$\times$1014 after 2$\times$ downsampling), ENeRF-Outdoor \cite{lin2022efficient} (960$\times$540, 2$\times$ downsampling), and D3DMV \cite{lin2021deep} (640$\times$360).
All three datasets provide multi-view sequences of dynamic actors in diverse static environments.
We use the same set of $K$ closest views to test all methods.
For testing, we use the \emph{Sear Steak} and \emph{Flame Steak} scenes from DyNeRF.
We do not perform any fine-tuning on the test datasets.
We observed sequences in the ENeRF dataset have incorrect synchronization, and inconsistent color calibration.
To mitigate the impact of these effects we compute all metrics at half the rendered resolution (480$\times$270) on this dataset.
\subsection{Evaluation Metrics}
\label{sec:evaluation-metrics}

We evaluate the quality of novel views using the standard PSNR, SSIM \cite{wang2004image}, LPIPS \cite{zhang2018perceptual}, and L1 metrics.
To evaluate temporal consistency, we use the spatio-temporal entropic difference (STED) \cite{soundararajan2012video}, and a version of temporal change consistency (TCC) \cite{zhang2019exploiting}, adapted to color images:
\begin{equation}
	\text{TCC} = \frac{1}{T-1} \sum_{t=1}^{T-1} \text{SSIM}(|\mathcal{I}^{t} - \mathcal{I}^{t+1}|, |I^{t} - I^{t+1}|)\text{,}
\end{equation}
where $\mathcal{I}^{t, t+1}$ and $I^{t, t+1}$ are the rendered and ground truth images respectively.
Finally, following \citet{long2021multi}, we measure the standard deviation of the L1 error across all frames (SDT) to evaluate temporal consistency, and across all views (SDV) for view consistency.

\subsection{Results}
\label{sec:results}
\setlength{\tabcolsep}{1.5mm}
\begin{table}
\small
\caption{\textbf{Quantitative ablation study.}  We evaluate our method on three key contributions: using a network to blend forward-rendered input views, using TSDF depth to guide the blending network, and using difference masks for temporal filtering.
\textit{No Blending} directly renders all input-view Gaussians into the target view in a single pass \cite{zheng2024gps}. \textit{Distance-based Blending} uses the distance of the target view from each input to determine the blending weights \cite{mildenhall2019local}.}
\vspace{-4mm}
\begin{center}
\begin{tabular}{lcccc}
\toprule
Variant & PSNR$\uparrow$ & LPIPS$\downarrow$ & TCC$\uparrow$ & STED$\downarrow$ \\
\midrule
No Blending             & 24.64  & 0.264 & 0.743 & 483.3\pad \\
Distance-based Blending & 26.08  & 0.211 & 0.858 & 235.3\pad\\
No TSDF Guidance        & 31.16 & 0.197 & 0.869 & \pad95.42 \\
No Temporal Filtering   & 31.16 & 0.195 & 0.869 & \pad89.62 \\
\midrule
Ours & \textbf{31.17} & \textbf{0.191} & \textbf{0.872} & \pad\textbf{68.76} \\
\bottomrule
\end{tabular}
\end{center}
\vspace{-3mm}
\label{table:ablation}
\end{table}

We evaluate view and temporal consistency in \cref{table:quantitative-vt-results} and \cref{fig:results-view-consistency}. We use epipolar plane images (EPI), and temporal slices of a fixed camera to visualize change across time and space respectively. Our results are sharper than IBRNet, show less flickering at boundaries than ENeRF, and suffer no abrupt changes like GPSG. Please refer to the supplemental videos for additional results on dynamic scenes.

A quantitative evaluation of the novel view rendering quality of all methods is presented in \cref{table:quantitative-results};
qualitative results are shown in \cref{fig:results-qualitative}. Our method reconstructs sharper edges, and finer details in both foreground and background. Even though the ENeRF-Outdoor dataset is challenging due to camera mis-synchronization, our method shows a marked improvement over the baselines, particularly at the edges of the actor and in disoccluded regions of the background. The latter are especially tough for ENeRF, which shows loss of detail at occlusion edges in all three datasets.

\subsection{Ablations}
\label{sec:ablations}
We ablate the core contributions of our work in \Cref{table:ablation} to quantitatively evaluate their impact. These include: 
\begin{enumerate*}[start=1, label=\arabic*)]
\item A network to blend and inpaint the forward-rendered views,
\item the use of a TSDF to guide the blending network, and
\item temporally filtering the input-view depths using difference masks.
\end{enumerate*}
To evaluate the effect of the TSDF depth, we retrain our network without the guidance depth $\mathcal{D}^t$ as input. We observe that our complete pipeline provides the best results, with the most significant gains in temporal consistency.

\cref{table:ablation_computation_time} shows the computation time of each stage of our method on the DyNeRF dataset.
We observe that even without TensorRT optimization, our total runtime ($\approx$100\,ms) is faster than the closest baseline GPSG, which runs in 143.4\,ms (\cref{fig:results-view-consistency}). All runtimes exclude disk read time.

\begin{table}[h!]
\footnotesize
\caption{Computation time of each stage in DyNeRF dataset.}
\label{table:ablation_computation_time}
\centering
\begin{tabular}{ccc|c}
\toprule
Forward-splatting & TSDF & Blending & Blending (w/o TensorRT) \\
\midrule
10\,ms    &6\,ms    &23\,ms & 65\,ms\\ 
\bottomrule
\end{tabular}
\label{table:runtime}
\vspace{3em}
\end{table}

\begin{figure}[t]
  \centering
  \includegraphics[width=0.9\linewidth,trim=0cm 0.3cm 0cm 0.0cm,clip]{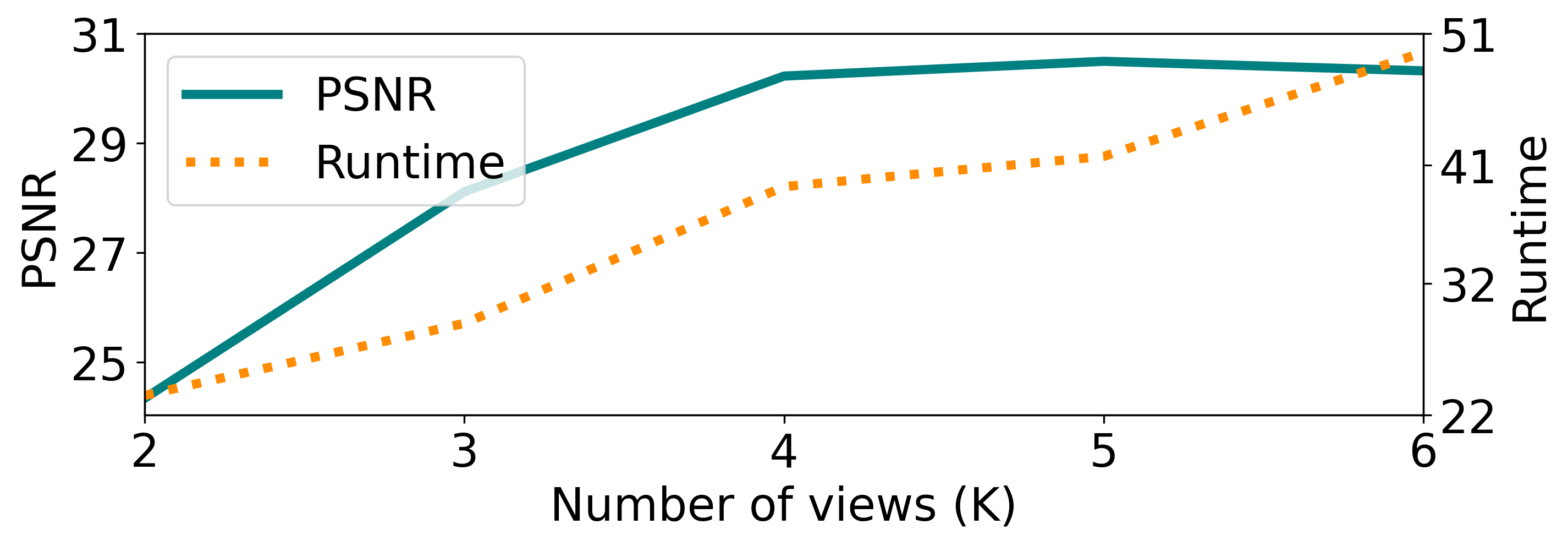}
   \caption{\textbf{Ablation study of the number of input views.} Runtime increases roughly linearly with the number of input views $K$. However, rendering quality saturates for $K \geqslant 4$ views.}
   \label{fig:input_num}
    \vspace{-1em}
\end{figure}

The choice to use a subset of $K$ cameras is motivated by efficiency, as each view adds to the TSDF, splatting, and blending time. We ablate the choice of $K$ in \cref{fig:input_num}.
\vspace{-1em}

\subsection{Limitations}
\label{sec:discussion}

\begin{figure}[h]
\centering
\includegraphics[trim=0cm 13.6cm 16cm 0cm, clip=true, width=1.0\columnwidth]{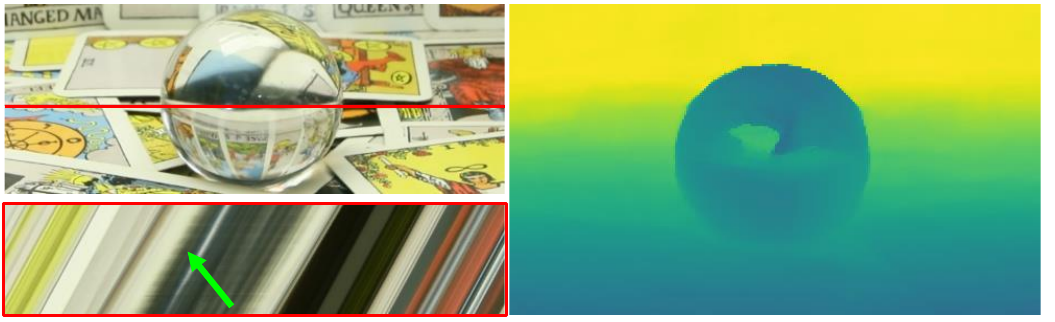}
\caption{\textbf{Limitations:} Inaccurate TSDF depth (right) can lead to view inconsistency in novel views, seen as EPI discontinuities. }
\label{fig:discussion}
\vspace{-1em}
\end{figure}

By using global geometry, our method promotes consistency in the rendered views. A limitation of this approach, however, is that it relies strongly on accurate geometry reconstruction. 
It fails, for instance, in specular regions where depth is usually bad~\cref{fig:discussion}. While the blending network still produces reasonable output for a single target in this case, the results suffer temporal and view consistency artifacts. Furthermore, our blending network requires well-synchronized and calibrated cameras for well-aligned forward projected input $\mathcal{I}_k^t$. Errors in camera calibration can lead to blurring artifact.

\section{Conclusion}
\label{sec:conclusion}
We present a novel method for consistent video synthesis in an online multi-view setting. 
This setting requires continuous and smoothly changing outputs in both the view and temporal domains, but it can easily suffer from inconsistency artifacts.
To address these challenges, our approach first fuses temporally filtered depth maps into a global structure to acquire view-temporal depth in the novel view. Then, this global depth information is used to guide the blending of forward-projected RGB views. By using multi-view temporally consistent depth to mediate the output of the blending network, our method produces consistent novel views. We show that our method effectively reconstructs a variety of scenes while running faster than previous work.

{
    \small
    \bibliographystyle{ieeenat_fullname}
    \bibliography{main}
}

\clearpage
\setcounter{page}{1}
\maketitlesupplementary
\renewcommand{\thesection}{A.\arabic{section}}
\setcounter{section}{0}

\section{Image-based TSDF}
We utilize an image-based TSDF \cite{lawrence2021project} that integrates depth along each ray on the fly from $K$ input depth maps $\{D_k,~k\!\in\!(1, ..., N)\}$. 
Specifically, for a world space point $\mathbf{p}\in\mathbb{R}^3$ along a viewing ray, we use the known camera intrinsics $\mathbf{K}$ and pose parameters $\mathbf{R}$, $\mathbf{t}$ of each view to project the point into the $K$ cameras as $(x_k, y_k, z_k) = \mathbf{K}_k(\mathbf{R}_k\mathbf{p} + \mathbf{t}_k)$. Then, we calculate the signed distance $s_{k}$ as the difference between the transformed point's $z$-value, and $D_k$ sampled at the normalized pixel locations $(u_k, v_k) = (x_k/z_k, y_k/z_k)$:
\begin{equation}
s_{k} = z_k - D_k\big[u_k, v_k\big]
\end{equation}
where the square brackets denote the sampling operation. We drop the superscript $t$ indicating time for convenience.

The truncated signed-distance value $s$ at point $\mathbf{p}$ is then computed by fusing $s_k$ from all input views as,
\begin{equation}
	\label{eq:tsdf_depth_render}
	s  =\sum_{k=1}^{K} \omega_{k} \cdot \text{clamp}(s_{k}, -\tau, \tau) \text{,}
\end{equation}
where $\tau \!=\! 0.02$\,m is the truncation threshold. The fusion weight $\omega_k$ handles noise in the input depth maps and is computed as the depth variation in a $w \times w$ pixel neighborhood $\mathcal{N}$ around the projected location $(u_k, v_k)$:
\begin{equation}
\omega_{k}=\min\!\Big(0.001 \cdot \Big( \frac{\nu_k}{w \times w} \Big)^{-1/2}, 1.0\Big) \text{,}
\end{equation}
\begin{equation}
\nu_{k}=\sum_{(p, q)\in\mathcal{N}}{\min\!\Big((D_{k}[u_k, v_k] - D_{k}[p, q])^{2}, \tau^{2}\Big)} \text{.}
\end{equation}
We use $w=7$ for our experiments and, following \citet{lawrence2021project}, set $\omega_{k} \!=\! 0$ if $s_{k} \!<\! -\tau$. We advance along the ray with a step size of $0.8s$ until a surface intersection is detected. This is indicated by a change of sign in the value of $s$. We subsequently perform three steps of a bisection search to refine the intersection depth.

\section{View and Temporally Consistent Depth}
We encourage temporal consistency in the TSDF depth map $\mathcal{D}^t$ by integrating the rendered depth $\mathcal{D}^{t-1}$ from the previous time frame, along with the input-view depth maps $D_k^t$. To account for dynamic regions we use the difference masks $M^{t}_{k}$ (Equation (1), main paper) to estimate fusion weights for $\mathcal{D}^{t-1}$. Specifically, we forward project all $K$ difference masks into the novel view using Gaussian splatting \cite{kerbl20233d}, and apply a channel-wise maximum to estimate the difference mask $\mathcal{M}^{t}$ in the novel-view.
We then fuse the signed distance computed from $\mathcal{D}^{t-1}$ into \cref{eq:tsdf_depth_render}, with the temporal fusion weight $\omega^t_\text{tmp}$ defined as: 
\begin{equation}	\omega^t_\text{tmp} = \min\left(\beta \cdot \omega^t_\text{acc}\cdot \max(1 \!-\! \mathcal{M}^{t}, 0), \eta\right) \! \text{, with}
\end{equation}
\begin{equation}
\omega_\text{acc}^t = \omega_\text{tmp}^{t-1} + \sum_k{\omega_k^{t-1}}\text{,}
\end{equation}
where $\eta \!=\! 15$ is the maximum fusion weight for $\mathcal{D}^{t-1}$, and $\beta$ controls the relative contribution of previous frames.

\section{Dense Pixel-sized 3D Gaussians}
As mentioned in Section 3.1 (main paper), we analytically compute the scale, rotation, color, and opacity parameters of dense per-pixel 3D Gaussians from each input view. 

This approach contrasts with previous work \cite{zheng2024gps,tu2024tele} that uses a neural network to predict per-pixel Gaussian parameters. In our experiments, we observed no advantage from using a network to predict parameters for the kind of dense Gaussian point cloud that we get from projecting a depth map to 3D. One possible motivation for using a network is that it can learn to inpaint disoclussions by adjusting the scale of background Gaussians. However, we found that this ability often comes at the cost of overall reconstruction PSNR. Thus,  we chose to address disocclusions using our geometry-guided blending network instead. 

\begin{figure}[t]
\centering
\includegraphics[trim=0cm 16cm 25cm 0cm, clip=true, width=1.0\columnwidth]{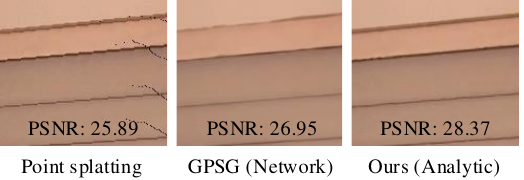}
\caption{\textbf{Forward rendering a single image.} Traditional point splatting suffers from aliasing artifacts such as empty pixels, and jagged edges. Network-based models seek to inpaint disocclusions by allowing the Gaussians' scale to vary. However, this affects overall PSNR. Our pixel-scaled Gaussians generate high-frequency details while avoiding aliasing.}
\label{fig:pixel-gaussians}
\end{figure}

Why use 3D Gaussians at all? We noticed that compared to traditional point splatting \cite{aliev2020neural}, Gaussians enable occlusion-handling and suffer from fewer aliasing artifacts, such as jagged edges and gaps between neighboring pixels \cref{fig:pixel-gaussians}. 

\section{Network Architecture}
Our blending network $\Theta(\cdot)$ is a four-layer U-Net \cite{ronneberger2015u}. The input to the network consist of,
\begin{enumerate}
\item The $K$ forward-rendered images $\{\mathcal{I}_k^t\} \in \mathbb{R}^{3K \times H\times W}$ , 
\item Depth maps $\{D_k^t\} \in \mathbb{R}^{K \times H\times W}$,
\item Alpha maps $\{\mathcal{\alpha
}_k^t\} \in \mathbb{R}^{K \times H\times W}$, 
\item The TSDF depth $\mathcal{D}^t \in \mathbb{R}^{1 \times H \times W}$. 
\end{enumerate}

\vspace{1mm}
\noindent
Additionally, we provide camera distance, and viewing angle information. Specifically, we use

\begin{enumerate}
\setcounter{enumi}{4} 
\item The dot product between each input-view ray direction, and the normals from the TSDF depth $\mathcal{D}^t$, $\in \mathbb{R}^{K \times H \times W}$
\item The dot product between each input-view ray direction, and the target view direction, $\in \mathbb{R}^{K\times H \times W}$. 
\item The distance between the input and target cameras, repeated spatially to get a map $\in \mathbb{R}^{K \times H \times W}$.
\item The dot product between the input and target viewing directions, repeated spatially to get a map $\in \mathbb{R}^{K \times H \times W}$.
\end{enumerate}

\vspace{1mm}
\noindent
In summary, the input to $\Theta(\cdot)$ is a tensor $\in \mathbb{R}^{(9K\!+\!1) \times H \times W}$

\section{Training Procedure}
We use the ScanNet \cite{dai2017scannet}, DyNeRF \cite{li2022neural}, and Google Spaces dataset \cite{flynn2019deepview} to train our network. ScanNet provides a dense views of general indoor scenes captured with a depth sensor.
We use three scenes from DyNeRF to account for large-baseline stereo scenarios, and utilize the remaining scenes for testing.
The Google Spaces dataset is used to provide training in small-baseline stereo settings. For the ScanNet dataset, we first select a novel time frame and then choose corresponding input frames from a range of $\pm$30 frames, excluding frames $[-4, 4]$ to avoid selecting the closest frames to the novel view.
In contrast, for the DyNeRF and Spaces datasets, we manually curate stereo pairs and generate depth maps for each view using RAFT-Stereo \cite{lipson2021raft}.
We then randomly choose input and novel views from at a fixed time frame for training.

\section{Testing Datasets}
The DyNeRF dataset \cite{li2022neural} consists of 18–22 cameras with a resolution of 2704$\times$2028 pixels, and features well-synchronized indoor scenes.
We use two scenes containing 22 cameras as the test set. We test all methods at half the original resolution (1352$\times$1014 pixels).

The ENeRF-outdoor dataset \cite{lin2022efficient} consists of 18 cameras capturing multiple actors in an outdoor setting. Each camera has a resolution of 1920$\times$1080 pixels.
Again, we use half the original resolution (960$\times$540) for testing all methods. Furthermore, we found that this dataset has imperfect camera synchronization, and suffers from color calibration errors. To mitigate the impact of these on quantitative metrics, we further downscale the rendered images to  480$\times$270 for the quantitative evaluation in Tables 1 and 2 of the main paper. 

The D3DMV dataset \cite{lin2021deep} includes 10 cameras capturing outdoor scenes. We use the compressed version of the datset with a resolution of 640$\times$360 pixels.

\section{Evaluation Procedure}
For all three datasets, we use the provided camera parameters. We evaluate the metrics for the first 100 frames. We select a subset of cameras as the test views. We use nine test views for DyNeRF, three test views for ENeRF, and four test views for the D3DMV dataset. We select the $K$ input cameras closest to the test view -- excluding the test view itself -- as inputs. We use $K \!=\! 4$ for DyNeRF and ENeRF-outdoor, and $K \!=\! 2$ for D3DMV in accordance with existing online multi-view methods \cite{lin2022efficient,lin2021deep,lawrence2021project,tu2024tele,zheng2024gps}.

\section{Additional Results}

\begin{figure}[h]
\centering
\adjustbox{margin=0pt 0pt 0pt 0pt}{
\includegraphics[width=0.6\linewidth,trim=0cm 13.5cm 23cm 0.0cm,clip]{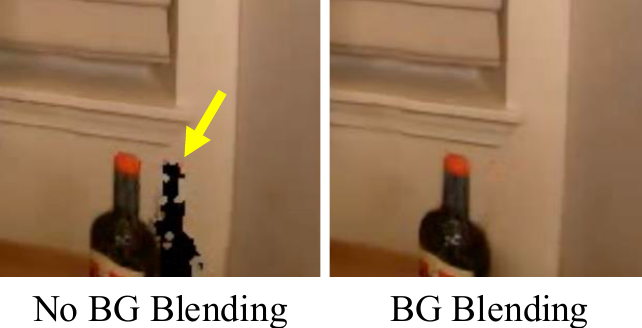}}
   \caption{\textbf{Impact of background blending.} Our background image blending successfully inpaints the empty regions in the forward-warped images caused by occluded areas in the input views.}
   \label{fig:ablation_bg}
\end{figure}

\paragraph{Background image blending.}
\Cref{fig:ablation_bg} shows the impact of background image blending in our network, allowing it to inpaint disocclusion holes in forward-warped images.

\begin{figure}[h]
  \centering
   \vspace{-0.3em} 
  \includegraphics[width=1.0\linewidth,trim=0cm 13.5cm 12cm 0.0cm,clip]{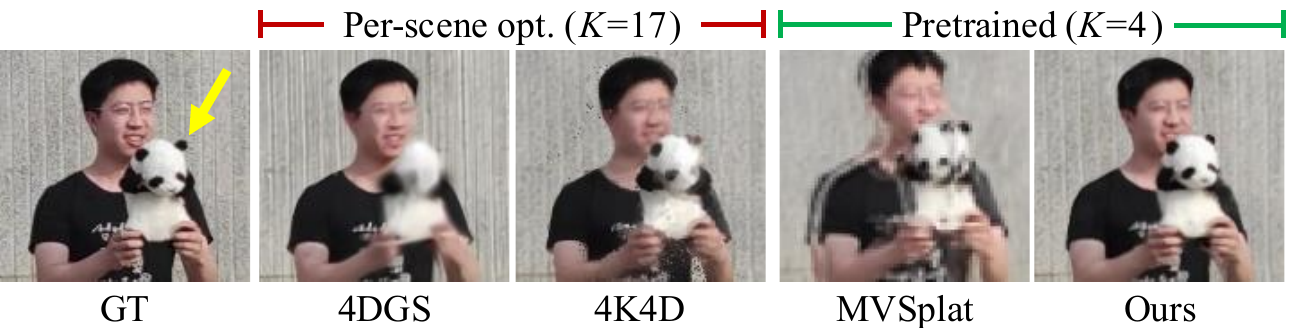}
  \vspace{-1.5em}
   \caption{\textbf{Qualitative comparison on the ENeRF-Outdoor~\cite{lin2022efficient} dataset.} The baseline methods produce blurry results despite per-scene optimization with dense views (4DGS, 4K4D) or a pretrained network with sparse views (MVSplat). Our method reconstructs fine details, including both moving humans and the background.}
   \vspace{-1em}
   \label{fig:additional_comp}
\end{figure}

\paragraph{Additional comparisons.} We additionally compare our method with offline dynamic scene reconstruction methods (4DGS \cite{wu20244d} and 4K4D \cite{xu20244k4d}), and sparse multi-view reconstruction method (MVSplat \cite{chen2024mvsplat}) in \cref{fig:additional_comp}. Even though 4K4D and 4DGS optimize per scene and use more views ($K\!=$17), our method shows better and sharper results.

\end{document}